\PassOptionsToPackage{normalem}{ulem}
\documentclass[]{arxiv_template}

\usepackage{url}
\usepackage{amssymb}
\usepackage{colortbl}
\usepackage{xspace}
\usepackage{makecell}
\usepackage{longtable}
\usepackage{listings}
\usepackage{fancyvrb}
\usepackage{textcomp}

\definecolor{darkblue}{rgb}{0, 0, 0.5}
\definecolor{lightgreen}{rgb}{0.9,1,0.9}
\definecolor{checkgreen}{rgb}{0,0.6,0}
\definecolor{todocolor}{rgb}{0.0, 0.2, 0.8}
\lstdefinestyle{filetree}{
  basicstyle=\small\ttfamily,
  frame=single,
  backgroundcolor=\color{white},
  rulecolor=\color{gray},
  escapeinside={(*@}{@*)},
  breaklines=true,
  columns=fullflexible,
}

\newcommand{\approach}{{SWE-Bench ProMax}\xspace}

\title{SWE-Bench ProMax: Benchmarking Agents on Large-Scale Multilingual Code Refactoring}

\makeatletter
\gdef\authorlist{%
  \begin{center}
    \authorfont\sffamily
    Yuling Shi$^{1*}$,
    Jinghan Xu$^{1*}$,
    Kelin Fu$^{2}$,
    Wenhao Zeng$^{1}$,
    Shilin He$^{4}$\\
    Lei Zhang$^{5}$,
    Yue Liu$^{6}$,
    Zelin Zhao$^{4}$,
    Terry Yue Zhuo$^{7}$,
    Jialun Cao$^{3}$,
    Siyu Ye$^{4}$\\
    Tianyu Liu$^{2}$,
    Kai Cai$^{4}$,
    Shing-Chi Cheung$^{3}$,
    Xiaodong Gu$^{1\dag}$
  \end{center}
}
\makeatother

\makeatletter
\gdef\affiliationlist{%
  \begin{center}
    \affiliationfont\sffamily
    $^{1}$Shanghai Jiao Tong University \quad
    $^{2}$Peking University\\
    $^{3}$The Hong Kong University of Science and Technology \quad
    $^{4}$Douyin Group\\
    $^{5}$University of Chinese Academy of Sciences \quad
    $^{6}$National University of Singapore\\
    $^{7}$Monash University
  \end{center}
}
\makeatother

\abstract{
As AI coding agents take on increasingly complex, long-horizon software engineering tasks, existing benchmarks are rapidly saturating and their evaluation quality has come under serious scrutiny: a recent audit found that nearly 60\% of unsolved SWE-bench Verified instances contain flawed tests---either overly narrow tests that reject correct solutions or overly broad tests that check unstated requirements---and that frontier models can verbatim reproduce gold patches from training data. Code refactoring, which requires coordinated, behavior-preserving changes across many files, offers a substantially harder and more realistic test of agent capability, yet remains underserved by current benchmarks. We introduce \approach, an expert-curated, multilingual code refactoring benchmark of 170 instances drawn from real commits across seven programming languages (Python, Java, TypeScript, Go, C, C++, and Rust). Every instance undergoes rigorous, multi-stage curation that directly addresses the quality problems identified in prior benchmarks: issue descriptions are rewritten from scratch to provide precise, unambiguous specifications, and test suites are manually reviewed to remove overly narrow and overly broad tests. Tasks with insufficient complexity or limited cross-file scope are filtered out, yielding a benchmark of challenging, large-scale refactoring tasks that average 11.5 modified files and 261.6 lines of code per instance, substantially exceeding the scale of existing benchmarks. Experiments with frontier models under two agent scaffolds show that the best model achieves only 41.2\% resolve rate, confirming that \approach presents a meaningful and unsaturated challenge for current AI coding agents. 
}

\checkdata[Dataset]{\url{https://huggingface.co/datasets/swe-bench-promax/SWE-Bench-ProMax}}
\correspondence{\email{xiaodong.gu@sjtu.edu.cn}}

\begin{document}

\maketitle
\tcbset{reset}

{\let\thefootnote\relax\footnotetext{$^*$Equal contribution. \quad $^\dag$Corresponding author.}}
{\let\thefootnote\relax\footnotetext{Published as a conference paper at COLM 2026.}}

\section{Introduction}
\label{sec:intro}

AI coding agents have advanced rapidly in recent years, with benchmarks playing a central role in measuring and driving progress~\citep{swe-bench,terminal-bench,shi2026earlyeval}. The field has moved from function-level evaluation, such as HumanEval~\citep{humaneval} and MBPP~\citep{mbpp}, to repository-level challenges like SWE-bench~\citep{swe-bench} and its successors~\citep{swe-bench-pro,multi-swe-bench,terminal-bench,swe-evo,huang2026deepswe}. As models improve, however, a pressing question emerges: do rising scores on these benchmarks genuinely reflect stronger software engineering ability, or are they reaching the limits of what current benchmark designs can measure?

Code refactoring, the disciplined process of restructuring existing code without changing its external behavior, is one of the most frequent activities in professional software development and a key mechanism for managing technical debt in evolving codebases~\citep{empirical-refactoring-study}. Industry leaders have identified refactoring as a prototypical long-horizon task: OpenAI highlights ``project-scale refactors'' as a primary use case for sustained, multi-context-window agents~\citep{gpt5-codex,gpt52-codex}, and Cursor reports that real-world developer tasks increasingly span many files and tools~\citep{cursorbench}. Yet as a benchmark domain, refactoring remains underexplored: existing benchmarks primarily target isolated bug fixes and feature implementations~\citep{swe-bench,swe-bench-pro}, while the few efforts focused on refactoring are limited in scale or restricted to a single language~\citep{swe-refactor,refactorbench}. We argue that real-world refactoring tasks, which require understanding large codebases~\citep{peng2025swe,shi2026codeocr}, coordinating changes across many files, and preserving subtle behavioral invariants, offer a substantially more demanding and realistic test of agent capability than the tasks that dominate current evaluations.

\begin{figure}[t]
    \centering
    \includegraphics[width=\linewidth]{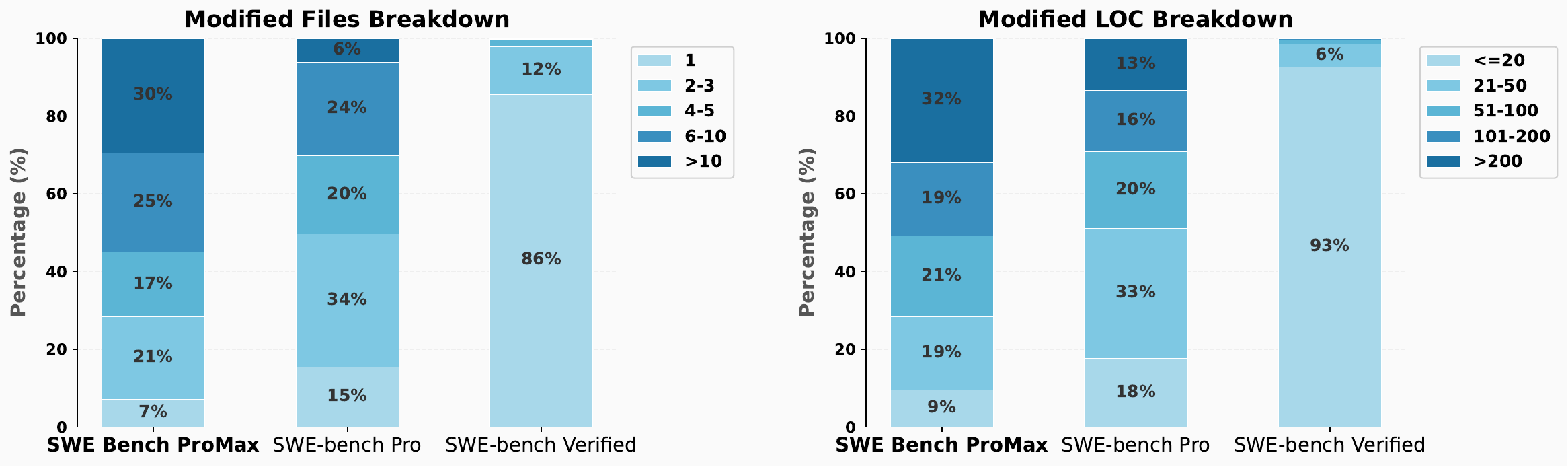}
    \caption{Distribution of modified files (left) and lines of code (right) per instance across benchmarks. \approach instances are substantially larger: 30\% modify more than 10 files and 32\% change over 200 lines of code, while 86\% of SWE-bench Verified instances modify only a single file.}
    \label{fig:breakdown}
\end{figure}

\begin{figure}[htbp]
\centering
\fbox{\begin{minipage}{0.95\linewidth}
\small
\textbf{Instance:} \texttt{nasa/fprime\#3422} \hfill \textbf{Language:} C++ \hfill \textbf{244 files} \hfill \textcolor{checkgreen}{\textbf{+591}} / \textcolor{black}{\textbf{$-$514}} lines

\vspace{4pt}
\hrule
\vspace{4pt}

\textbf{Modified files (excerpt):}

\ttfamily
\begin{tabular}{@{}l@{}}
Autocoders/Python/src/.../component/cpp.tmpl \\
Drv/BlockDriver/BlockDriverImpl.cpp \\
Fw/FPrimeBasicTypes.hpp \hfill \textcolor{checkgreen}{\textrm{\textbf{[NEW]}}} \\
Fw/FPrimeBasicTypes.h \hfill \textcolor{checkgreen}{\textrm{\textbf{[NEW]}}} \\
Fw/Types/BasicTypes.h \\
FppTest/component/active/ActiveTest.cpp \\
Os/Baremetal/TaskRunner/TaskRunner.cpp \\
Svc/ActiveLogger/ActiveLoggerImpl.cpp \\
\textcolor{gray}{\textrm{\textit{... and 238 more files}}} \\
\end{tabular}
\end{minipage}}
\caption{A representative \approach instance: refactoring NASA's F'Prime flight software framework requires coordinated changes across 244 files spanning the entire codebase.}
\label{fig:intro-example}
\end{figure}

Building a reliable refactoring benchmark, however, exposes two weaknesses that pervade existing benchmarks more broadly. The first is difficulty. On SWE-bench Verified, frontier agents now exceed 75\% resolve rate~\citep{gpt5}, and the gap among top systems continues to narrow---prompting both academic~\citep{metr-time-horizons} and industry~\citep{cursorbench} observers to question whether public benchmarks still meaningfully differentiate frontier capabilities. A key reason is scope: most existing benchmark instances involve modifications to a small number of files with limited lines of code. Real-world refactoring operates at a fundamentally different scale, often requiring coordinated changes across dozens of files and hundreds of lines of code, demanding sustained cross-file reasoning that current agents struggle to maintain~\citep{longcli-bench,desai2026swe}.
The second weakness is evaluation quality. Test suites in existing benchmarks suffer from two complementary defects: \emph{overly narrow} tests that enforce specific implementation details and reject functionally correct solutions, and \emph{overly broad} tests that check behavior not specified in the task description~\citep{swe-bench-pro,terminal-bench}. A recent audit of SWE-bench Verified found these defects in nearly 60\% of unsolved instances---35.5\% had narrow tests and 18.8\% had broad tests---leading OpenAI to deprecate the benchmark entirely~\citep{openai-swebench-deprecated}. Compounding this, problem descriptions are often imprecise or ambiguous, and growing evidence of data contamination in benchmarks sourced from public repositories further erodes the correlation between benchmark scores and genuine capability~\citep{openai-swebench-deprecated,shi2024between}. Additionally, despite growing multilingual efforts~\citep{multi-swe-bench,swe-polybench}, most benchmarks remain Python-centric, leaving open how well agents generalize across language paradigms such as Rust's ownership model, C's manual memory management, or Java's type hierarchies.

We introduce \approach to address these gaps. The benchmark comprises 170 refactoring instances drawn from real commits in actively maintained GitHub repositories, spanning seven programming languages: Python, Java, TypeScript, Go, C, C++, and Rust. As Figure~\ref{fig:breakdown} illustrates, \approach operates at a fundamentally different scale than existing benchmarks: 30\% of our instances modify more than 10 files and 32\% require over 200 lines of code, compared to SWE-bench Verified where 86\% of instances modify only a single file.
To make this concrete, Figure~\ref{fig:intro-example} shows a representative task from \approach: refactoring NASA's F'Prime flight software framework requires migrating \textcolor{black}{244 files} from a monolithic header to a new unified entry point---touching autocoders, drivers, OS layers, services, and build configurations---while preserving identical runtime behavior\footnote{\url{https://github.com/nasa/fprime/pull/3422}}. The full problem statement and additional examples across all seven languages are presented in Appendix~\ref{app:cases}.

Our contributions are as follows.
First, we ensure benchmark quality through rigorous, multi-stage expert curation. Issue descriptions are rewritten from scratch to provide precise, unambiguous specifications that fully define the expected refactoring. Test suites are manually reviewed to eliminate overly narrow and overly broad tests, and each problem statement is verified to serve as both a necessary and sufficient condition for the gold patch.
Second, the benchmark presents a genuine challenge to frontier models. During curation, tasks with insufficient complexity, limited cross-file scope, or too few lines of code are systematically filtered out. The remaining instances require large-scale, coordinated modifications averaging 11.5 files and 261.6 lines of code, and \textcolor{black}{the best model in our evaluation achieves only 41.2\% resolve rate. Notably, higher cost does not guarantee proportionally better performance: Claude Sonnet 4.6 averages \$4.77 per instance yet resolves 38.8\%, while GLM-5 attains a comparable 36.5\% at only \$0.24---suggesting that open-weight models can approach frontier performance at a fraction of the cost.} Agent trajectory analysis further reveals that failed attempts consistently modify fewer files than the gold patch requires while consuming more interaction rounds, pointing to incomplete cross-file coordination as the dominant failure mode.
Third, the benchmark spans seven diverse programming languages, enabling analysis of how language-specific features, including type systems, memory models, and build ecosystems, affect agent performance on code transformation tasks.

\section{Related work}
\label{sec:related}

\subsection{Coding benchmarks}

As frontier code models~\citep{deepseek-coder-v2,qwen3-coder,glm5,kimi-k2,gpt5} and autonomous agents~\citep{swe-agent,openhands,agentless,zhao2026immersion} have rapidly advanced, \textcolor{black}{with complementary work on efficient code reasoning, repository-level context construction~\citep{zhang2026fastcontext,hu2026line,agentless}, collaborative inference, and long-horizon agent memory~\citep{zeng2025pruning,wang2026swe,shi2025longcodezip,zeng2026glimprouter,li2025swe,gao2026swe,chen2025swe},} benchmarks have become the critical bottleneck for measuring real progress. Table~\ref{tab:benchmark-comparison} compares \approach with existing benchmarks across six dimensions.

\paragraph{From function-level to long-horizon evaluation.}
Early benchmarks such as HumanEval~\citep{humaneval}, MBPP~\citep{mbpp}, and LiveCodeBench~\citep{livecodebench} evaluate isolated code generation. SWE-bench~\citep{swe-bench} introduced repository-level evaluation from real GitHub issues, with subsequent work extending along multiple axes: Multi-SWE-bench~\citep{multi-swe-bench} and SWE-PolyBench~\citep{swe-polybench} broaden language coverage; SWE-bench Pro~\citep{swe-bench-pro} and SWE-EVO~\citep{swe-evo} target harder, long-horizon tasks (SWE-EVO averages 21 modified files); Terminal-Bench~\citep{terminal-bench} curates 89 hard CLI challenges with human-written verification; and \textcolor{black}{SWE-bench Live~\citep{swe-bench-live}, SWE-rebench~\citep{swe-rebench}, SWE-Factory~\citep{swe-factory} and Dockerless~\citep{zeng2026dockerless} address contamination, infrastructure and verification}. Current benchmarks are rapidly saturating---frontier agents exceed 75\% on SWE-bench Verified~\citep{swe-bench}---and recent studies show that agent capabilities on short tasks are doubling every few months~\citep{metr-time-horizons}, while all state-of-the-art agents achieve less than 20\% on long-horizon tasks~\citep{longcli-bench}. \approach targets this difficulty frontier: our instances require coordinated changes across 11.5 files and 261.6 lines of code on average, and \textcolor{black}{the best evaluated model achieves only 41.2\% resolve rate}.

\paragraph{Refactoring benchmarks.}
Despite the prevalence of refactoring in professional development, only two benchmarks specifically target it. RefactorBench~\citep{refactorbench} provides 100 handcrafted multi-file tasks across 9 Python repositories but is limited to a single language, with instances averaging only 4.3 modified files. SWE-Refactor~\citep{swe-refactor} offers 1,099 instances from 18 Java repositories with automated validation but covers only Java and lacks human verification of test quality. Neither combines multilingual coverage, large-scale complexity, and expert-curated test suites---the combination that \approach provides.

\paragraph{Expert curation.}
The degree of human curation varies widely across benchmarks (Table~\ref{tab:benchmark-comparison}). At the repository level, curation ranges from fully automated (SWE-bench~\citep{swe-bench}) to extensive: Multi-SWE-bench~\citep{multi-swe-bench} employs 68 annotators, Terminal-Bench~\citep{terminal-bench} invests approximately three reviewer-hours per task, and SWE-bench Pro~\citep{swe-bench-pro} rewrites issue descriptions and reviews test scope. However, even well-curated benchmarks retain quality issues: an audit of SWE-bench Verified found that nearly 60\% of unsolved instances had material defects in test design or problem descriptions~\citep{openai-swebench-deprecated}. \approach applies expert curation systematically to every instance: issue descriptions are rewritten from scratch as precise specifications informed by the gold patch and test suite, and inappropriate tests (overly narrow or overly broad) are identified and removed, ensuring mutual alignment between specification and evaluation.

\subsection{Code refactoring}

Traditional refactoring tools such as RefactoringMiner~\citep{refactoringminer} focus on \emph{detecting} refactorings from commit histories rather than generating them, and empirical studies reveal that 61 out of 100 identified refactoring types remain unsupported by existing engines~\citep{empirical-refactoring-study,automated-refactoring-discovery}. Recent work has begun evaluating LLMs as refactoring agents, revealing consistent limitations: agents perform only low-level edits and fail to address high-level design issues~\citep{agentic-refactoring}, struggle with complex context-dependent refactoring~\citep{refactoring-llm-empirical}, and achieve only 7.7\% alignment when autonomously discovering needed changes~\citep{codetaste}. Crucially, each study uses different evaluation methodologies---code smell counts~\citep{refactoring-llm-empirical}, compilability~\citep{refactoring-llm-fewshot}, alignment scores~\citep{codetaste}---making cross-study comparison impossible and motivating a standardized, execution-based benchmark with human-verified tests. \approach fills this role.

\section{\approach}
\label{sec:benchmark}

\begin{table}[t]
\centering
\small
\newcommand{\ch}{\textcolor{checkgreen}{\checkmark}}
\newcommand{\hd}[2]{\makecell{\textbf{#1}\\\textbf{#2}}}
\resizebox{\linewidth}{!}{
\begin{tabular}{lcccccc}
\toprule
\textbf{Benchmark} & \hd{Execution}{Based} & \hd{Repo}{Level} & \hd{Multi}{Lingual} & \hd{Refac-}{toring} & \hd{Avg. $>$5}{Files} & \hd{Expert}{Curated} \\
\midrule
HumanEval~\citep{humaneval} & \ch & & & & & \ch \\
MBPP~\citep{mbpp} & \ch & & & & & \ch \\
LiveCodeBench~\citep{livecodebench} & \ch & & & & & \\
\midrule
SWE-bench~\citep{swe-bench} & \ch & \ch & & & & \\
Multi-SWE-bench~\citep{multi-swe-bench} & \ch & \ch & \ch & & & \ch \\
SWE-PolyBench~\citep{swe-polybench} & \ch & \ch & \ch & & & \\
SWE-bench Pro~\citep{swe-bench-pro} & \ch & \ch & & & & \ch \\
SWE-EVO~\citep{swe-evo} & \ch & \ch & & & \ch & \\
Terminal-Bench~\citep{terminal-bench} & \ch & \ch & & & & \ch \\
\midrule
RefactorBench~\citep{refactorbench} & \ch & \ch & & \ch & & \ch \\
SWE-Refactor~\citep{swe-refactor} & \ch & \ch & & \ch & & \\
\midrule
\rowcolor{lightgreen}
\textbf{\approach (Ours)} & \ch & \ch & \ch & \ch & \ch & \ch \\
\bottomrule
\end{tabular}
}

\caption{Comparison of \approach with existing benchmarks. \textbf{Avg.\ $>$5 Files}: the average gold patch modifies more than 5 files. \textbf{Expert Curated}: problem descriptions and/or test suites undergo manual expert review or authoring beyond automated collection.}
\label{tab:benchmark-comparison}
\end{table}

This section describes how the benchmark is constructed, verified, and composed.

\subsection{Task formulation}

Each instance in \approach consists of four components: (1) a pre-configured Docker environment containing the target repository at the     commit immediately before the refactoring, with all dependencies installed; (2) an issue description specifying the intended refactoring in precise natural language; (3) a test suite that validates whether the refactoring has been correctly applied; and (4) a gold patch recording the original developer's solution.
Given the environment and the issue description, an agent must autonomously modify the repository so that all tests pass. An instance is considered \emph{resolved} if and only if the agent's modifications pass every test in the suite. This formulation is outcome-driven: we evaluate the final state of the repository rather than the specific commands or intermediate steps the agent takes.

\subsection{Dataset construction and curation}

\begin{figure}[t]
    \centering
    \includegraphics[width=0.9\linewidth]{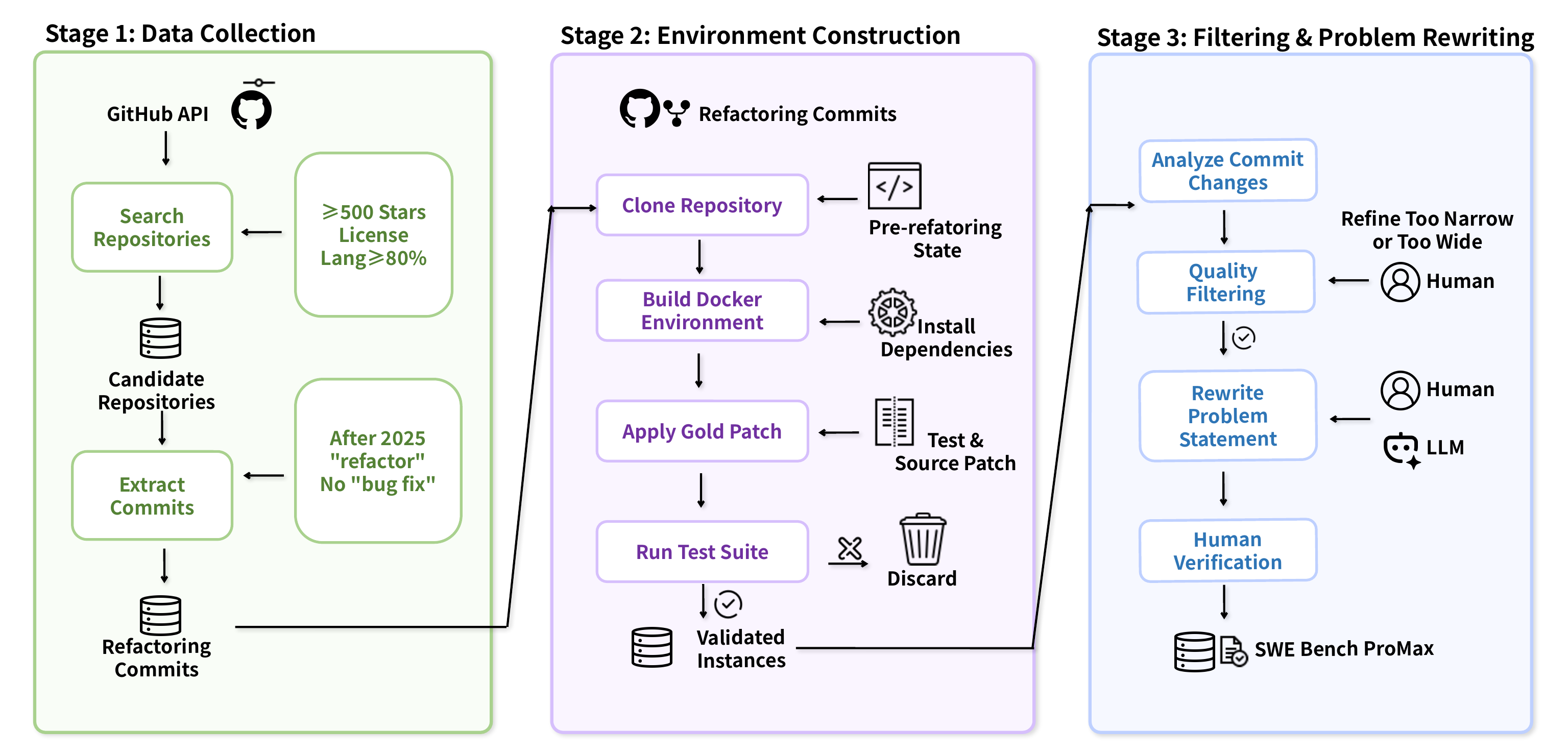}
    \caption{Data collection and curation pipeline for \approach.}
    \label{fig:data_collection}
\end{figure}

Figure~\ref{fig:data_collection} illustrates our three-stage pipeline, which progresses from automated collection through environment validation to expert-driven curation. The quality of a benchmark is determined not by its size but by the reliability of its evaluation~\citep{terminal-bench,swe-evo}; accordingly, every instance in \approach undergoes multi-stage review before inclusion, and our final benchmark retains only 170 out of 29,782 initial candidates.

\paragraph{Stage 1: Data collection.}
We use the GitHub API to identify candidate repositories that meet three criteria: at least 500 stars, an approved open-source license, and a primary language (comprising at least 80\% of the codebase) among our seven target languages. From these repositories, we extract commits submitted after January 2025 whose messages contain the keyword ``refactor'' but not ``bug fix,'' and that modify both test and non-test files. This yields a large initial pool of refactoring-related commits across all seven languages.

\paragraph{Stage 2: Environment construction.}
For each candidate commit, we construct an isolated Docker environment containing the repository at the pre-refactoring state with all build dependencies installed, leveraging automated environment construction tools~\citep{swe-factory,fu2026davinci} and SpecBuild~\citep{shi2026speculativeforkingturningidle}. The repository is cloned at the pre-refactoring commit, the gold patch (comprising both source and test changes) is applied, and the full test suite is executed. Instances where a working environment cannot be established or where the gold patch fails to pass the test suite are discarded, yielding a set of validated instances.

\paragraph{Stage 3: Filtering and problem rewriting.}
Because our instances are mined from commits rather than curated issues, the raw data lacks the precise problem descriptions that a benchmark requires---commit messages are written for fellow developers, not as task specifications for AI agents. The final stage therefore involves human experts working with LLM assistance to transform validated instances into high-quality benchmark tasks through four steps.

\emph{(1) Commit analysis.} Experts analyze the commit diff, assisted by LLMs that summarize changes and identify affected components, to understand the scope, intent, and structural impact of each refactoring before any filtering or rewriting decisions are made.

\emph{(2) Quality filtering.} Instances with insufficient complexity are removed: we discard tasks confined to a single file, those with too few lines of code modified, or those involving overly simplistic patterns. We also review each test suite to identify and remove \emph{overly narrow tests} that enforce specific implementation details rather than behavioral outcomes, and \emph{overly broad tests} that check behavior beyond the scope of the refactoring---directly addressing the quality defects that have undermined prior benchmarks~\citep{openai-swebench-deprecated}.

\emph{(3) Problem statement rewriting.} Original commit messages are typically terse (e.g., ``refactor auth module''), ambiguous, or reference internal context unavailable to an agent. Experts therefore rewrite the issue description from scratch with LLM assistance, producing a precise, self-contained specification that states which components should change, what the expected transformation is, and what behavioral invariants must be preserved. Each description is verified to serve as both a necessary and sufficient condition for the gold patch: a correct solution should satisfy the description, and the description should not admit unintended solutions.

\emph{(4) Human verification.} A final round of expert review ensures consistency across the issue description, test suite, and gold patch. The description must fully specify the refactoring, the tests must pass if and only if the refactoring is correctly applied, and no unstated requirements remain.

\subsection{Dataset composition}

\begin{figure}[t]
    \centering
    \begin{minipage}[c]{0.48\linewidth}
        \centering
        \small
        \begin{tabular}{lcc}
            \toprule
            & \textbf{Mean} & \textbf{Max} \\
            \midrule
            \textbf{Issue Description} & & \\
            \quad Tokens & 685.3 & 2,092 \\
            \midrule    
            \textbf{Gold Patch (source)} & & \\
            \quad \# Files & 11.5 & 182 \\
            \quad Lines of code & 261.6 & 4,503 \\
            \quad Tokens & 8,179.5 & 72,623 \\
            \midrule
            \textbf{Test Patch} & & \\
            \quad \# Files & 4.5 & 66 \\
            \quad Lines of code & 185.5 & 1,959 \\
            \quad Tokens & 3,980.8 & 52,031 \\
            \midrule
            \textbf{Total} & & \\
            \quad \# Files & 15.9 & 244 \\
            \bottomrule
        \end{tabular}
        \captionof{table}{Dataset statistics of \approach.}
        \label{tab:dataset-stats}
    \end{minipage}
    \hfill
    \begin{minipage}[c]{0.48\linewidth}
        \centering
        \includegraphics[width=\linewidth]{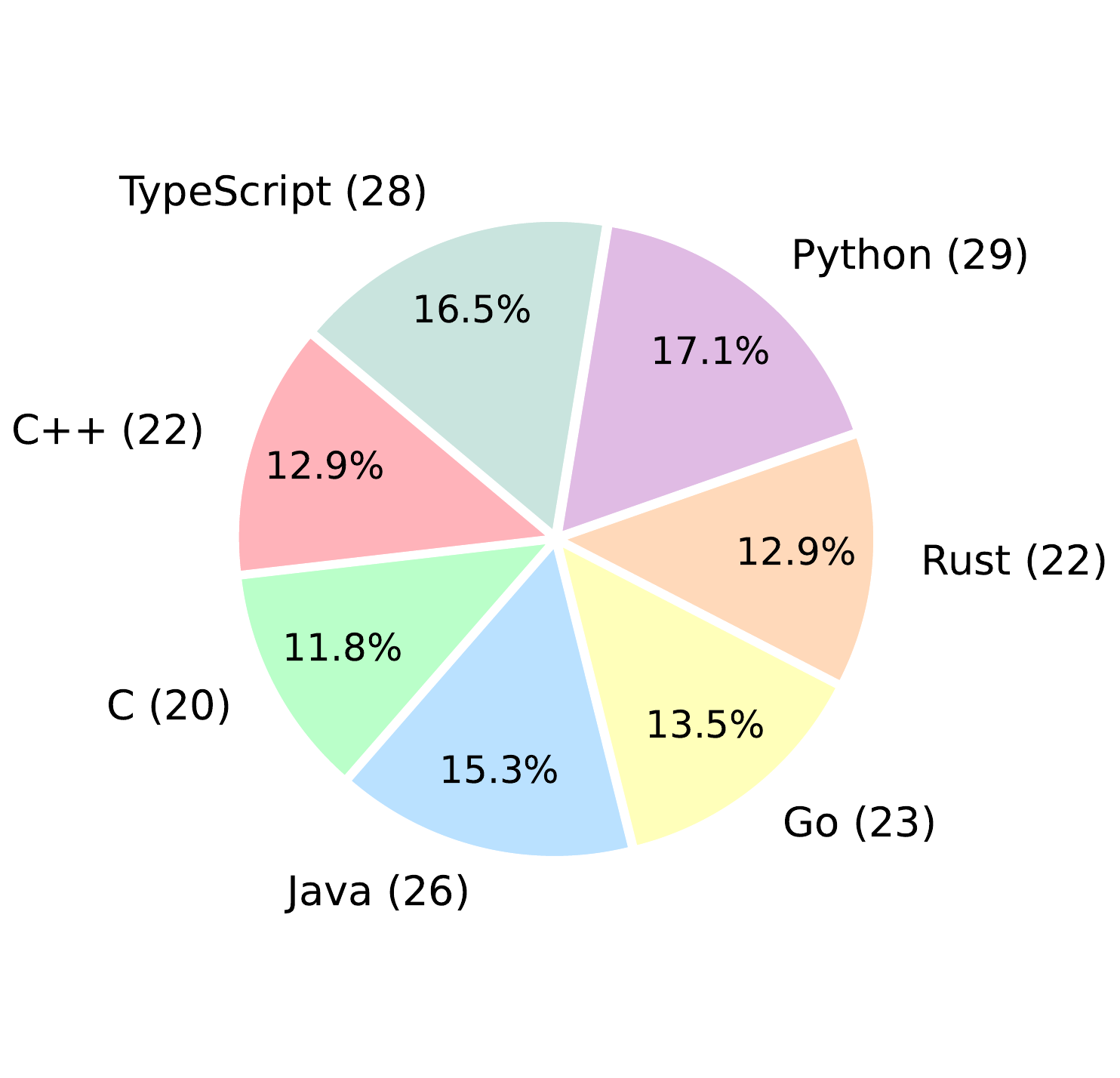}
        \caption{Language distribution in \approach.}
        \label{fig:language-distribution}
    \end{minipage}
\end{figure}

\paragraph{Language coverage.}
\approach spans seven programming languages that represent diverse paradigms: dynamically typed (Python), statically typed with garbage collection (Java, Go), gradually typed (TypeScript), systems languages with manual memory management (C, C++), and a language with an ownership-based memory model (Rust). Instances are drawn from 70 distinct repositories (Appendix~\ref{app:data}), ensuring broad coverage beyond a small number of projects. This diversity enables us to examine how language-specific features affect agent performance on refactoring tasks.

\paragraph{Scale and complexity.}
A defining characteristic of \approach is the scale of required code changes. As shown in Table~\ref{tab:dataset-stats}, gold patches average 11.5 source files and 261.6 lines of code (\textcolor{black}{8,179.5 tokens}), with the most complex instances modifying \textcolor{black}{up to 182 source files}. Test patches add a further 4.5 files and 185.5 lines on average, bringing the total to 15.9 files per instance. Issue descriptions average 685.3 tokens, providing detailed specifications of the required refactoring. As shown in Figure~\ref{fig:breakdown}, our instances require substantially more extensive modifications than those in SWE-bench Verified or SWE-bench Pro, reflecting the true complexity of production-scale refactoring.

\section{Experiments}
\label{sec:exp}

\subsection{Experimental setup}

\paragraph{Agent scaffolds.}
\textcolor{black}{We evaluate all models under two agent scaffolds. The first is mini-swe-agent~\citep{swe-agent}, a minimal reimplementation of the widely adopted SWE-agent scaffold that has been used as the default evaluation framework~\citep{swe-bench-pro,multi-swe-bench,terminal-bench}; it provides file viewing, editing, searching, and bash execution in an iterative observe-think-act loop. The second is OpenHands~\citep{openhands}, an open platform for generalist software agents that equips models with a richer runtime, including sandboxed command execution and structured file-editing tools. Given the large scale of refactoring patches in \approach, we set a step limit of 300 and a cost limit of \$10 per instance under both scaffolds. Applying the same scaffolds and limits to all models ensures fair comparison.}

\paragraph{Execution environment.}
Each instance ships with a pre-built, isolated Docker container constructed during benchmark curation using SWE-Factory~\citep{swe-factory} and SpecBuild~\citep{shi2026speculativeforkingturningidle} and manually verified to be functional. The container contains the repository at the pre-refactoring commit with all dependencies installed, so evaluation requires no additional environment setup.

\paragraph{Evaluation protocol.}
An instance is \emph{resolved} if the agent's modifications pass every test in the suite. Our primary metric is the \textbf{resolve rate} (Pass@1): the percentage of resolved instances. We evaluate on all 170 instances and report both overall and per-language resolve rates. We also analyze average cost (\$) per instance across models.

\subsection{Models}

We evaluate six frontier models spanning both proprietary and open-weight families:

\textbf{Proprietary models.}
(1)~Gemini-3-Pro~\citep{gemini3pro}, Google's flagship model;
(2)~Claude Sonnet 4.6~\citep{claudesonnet46}, Anthropic's latest Sonnet-class model;
(3)~GPT-5.2~\citep{gpt52}, OpenAI's most recent reasoning model.

\textbf{Open-weight models.}
(4)~GLM-5~\citep{glm5}, a mixture-of-experts model from Zhipu AI;
(5)~Kimi-K2.5~\citep{kimi-k25}, Moonshot AI's agentic intelligence model;
(6)~Qwen3.5~\citep{qwen3-coder}, Alibaba's latest code-oriented model.

This selection covers diverse architectures, scales, and training paradigms, enabling analysis of how these factors affect refactoring performance.

\section{Results and analysis}
\label{sec:results}

\subsection{Main results}
\begin{table}[t]
    \centering
    \small
    \resizebox{\linewidth}{!}{
    \begin{tabular}{lcccccccccc}
        \toprule
        & \textbf{Overall} & \textbf{Avg. Steps} & \textbf{Avg. Cost} & \multicolumn{7}{c}{\textbf{Per-Language Resolve Rate}} \\
        \cmidrule(lr){5-11}
        & & & & \textbf{Py} & \textbf{Java} & \textbf{TS} & \textbf{Go} & \textbf{C} & \textbf{C++} & \textbf{Rust} \\
        \midrule
        \rowcolor[gray]{0.93}
        \multicolumn{11}{l}{\textbf{Mini-SWE-Agent}} \\
        \rowcolor[gray]{0.88}
        \multicolumn{11}{l}{\textit{Proprietary}} \\
        Gemini-3-Pro & 26.5 & 58.0 & \$0.60 & 17.2 & 15.4 & 14.3 & 26.1 & 50.0 & 45.5 & 27.3 \\
        Claude Sonnet 4.6 & 30.6 & 99.5 & \$2.32 & 13.8 & 19.2 & 32.1 & 13.0 & 65.0 & 40.9 & 40.9 \\
        GPT-5.2 & 21.8 & 25.2 & \$0.19 & 17.2 & 15.4 & 21.4 & 13.0 & 45.0 & 31.8 & 13.6 \\
        \midrule
        \rowcolor[gray]{0.88}
        \multicolumn{11}{l}{\textit{Open-weight}} \\
        GLM-5 & 22.9 & 108.9 & \$0.10 & 13.8 & 7.7 & 25.0 & 21.7 & 50.0 & 22.7 & 27.3 \\
        Kimi-K2.5 & 26.5 & 85.3 & \$0.37 & 17.2 & 23.1 & 21.4 & 17.4 & 60.0 & 31.8 & 22.7 \\
        Qwen3.5 & 20.6 & 155.4 & \$0.93 & 17.2 & 7.7 & 10.7 & 13.0 & 45.0 & 27.3 & 31.8 \\
        \midrule
        \rowcolor[gray]{0.93}
        \multicolumn{11}{l}{\textbf{OpenHands}} \\
        \rowcolor[gray]{0.88}
        \multicolumn{11}{l}{\textit{Proprietary}} \\
        Gemini-3-Pro & 19.4 & 51.2 & \$1.49 & 13.8 & 19.2 & 0.0 & 8.7 & 45.0 & 36.4 & 22.7 \\
        Claude Sonnet 4.6 & 38.8 & 117.9 & \$4.77 & 17.2 & 30.8 & 53.6 & 26.1 & 50.0 & 36.4 & 63.6 \\
        GPT-5.2 & 41.2 & 115.1 & \$3.60 & 48.3 & 19.2 & 35.7 & 26.1 & 75.0 & 36.4 & 54.5 \\
        \midrule
        \rowcolor[gray]{0.88}
        \multicolumn{11}{l}{\textit{Open-weight}} \\
        GLM-5 & 36.5 & 114.2 & \$0.24 & 20.7 & 34.6 & 28.6 & 34.8 & 65.0 & 45.5 & 36.4 \\
        Kimi-K2.5 & 32.9 & 99.6 & \$0.72 & 24.1 & 30.8 & 10.7 & 43.5 & 70.0 & 45.5 & 18.2 \\
        Qwen3.5 & 36.5 & 141.2 & \$0.78 & 37.9 & 26.9 & 17.9 & 39.1 & 65.0 & 54.5 & 22.7 \\
        \bottomrule
    \end{tabular}
    }
    \caption{Resolve rates (\%) of all evaluated models across scaffolds on \approach.}
    \label{tab:main-results}
\end{table}

\textbf{All models find \approach challenging.} The best-performing model, GPT-5.2, achieves only 41.2\% resolve rate---far below the 75\%+ that frontier agents achieve on SWE-bench Verified~\citep{gpt5}. This confirms that multi-file refactoring, with its requirements for cross-file coordination and behavioral preservation, remains a substantial unsolved challenge.

\textcolor{black}{\textbf{Open-weight models are competitive with proprietary ones at a fraction of the cost.} Under OpenHands, GLM-5 and Qwen3.5 (both 36.5\%) and Kimi-K2.5 (32.9\%) come within a few points of GPT-5.2 (41.2\%) and Claude Sonnet 4.6 (38.8\%), while spending only a fraction as much per instance (\$0.24, \$0.78, and \$0.72 versus \$3.60 and \$4.77). The choice of scaffold also matters substantially: every model except Gemini-3-Pro improves markedly when moving from mini-swe-agent to OpenHands (e.g., GPT-5.2 from 21.8\% to 41.2\%), suggesting that richer runtime tooling is particularly beneficial for large-scale refactoring tasks.}

\textbf{Performance varies substantially across languages.} No single model dominates all languages. \textcolor{black}{Claude Sonnet 4.6 leads on TypeScript (53.6\%) and Rust (63.6\%), GLM-5 leads on Java (34.6\%), while GPT-5.2 performs best on Python (48.3\%) and C (75.0\%). Kimi-K2.5 achieves its best result on Go (43.5\%), while Qwen3.5 performs best on C++ (54.5\%).} This diversity suggests that different model architectures and training data compositions lead to complementary language-level strengths.


\textcolor{black}{\textbf{TypeScript and Rust show surprising variance.} Despite their reputation as complex languages, TypeScript and Rust yield high resolve rates for some models (Claude Sonnet 4.6: 53.6\% on TypeScript and 63.6\% on Rust; GPT-5.2: 54.5\% on Rust) while remaining difficult for others (Gemini-3-Pro: 0.0\% on TypeScript; Kimi-K2.5: 18.2\% on Rust). This variance may reflect differences in language-specific training data rather than inherent language difficulty.}

\subsection{Agent behavior analysis}

\begin{figure*}[t]
    \centering
    \includegraphics[width=\textwidth]{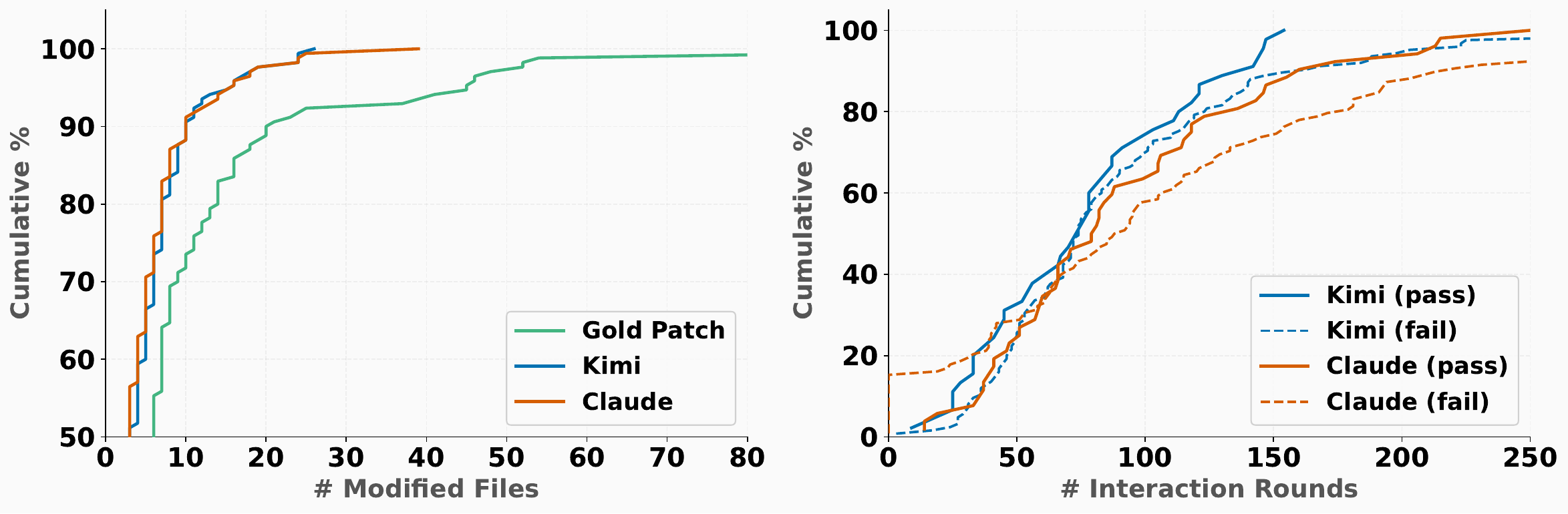}
    \caption{\textbf{Left:} Cumulative distribution of files modified by agents (Claude Sonnet 4.6, Kimi-K2.5) versus the gold patch. Agents modify fewer files than required overall, with the gap widening for larger patches. \textbf{Right:} Cumulative distribution of interaction rounds on resolved (solid) versus unresolved (dashed) instances. Failed attempts consume substantially more rounds than successful ones.}
    \label{fig:failure-analysis}
\end{figure*}

Figure~\ref{fig:failure-analysis} reveals two complementary failure patterns across \textcolor{black}{two representative models (Claude Sonnet 4.6 and Kimi-K2.5)}.

The left panel compares the number of files modified by each agent against the gold patch. Both Claude Sonnet 4.6 and Kimi-K2.5 closely track the gold patch distribution for small changes (up to ${\sim}$5 files), but diverge sharply for larger patches: whereas the gold patch CDF reaches 90\% only around 20 files, both agents reach 90\% by approximately 10 files. This indicates that the dominant failure mode is \emph{incomplete refactoring}---agents identify and modify some affected files but fail to propagate changes to all locations requiring coordinated updates. Critically, this is not a matter of failing to find the right files: agents often correctly locate and edit the core files involved in a refactoring, but stop short of applying the same transformation to peripheral call sites, documentation, configuration files, and test fixtures that also need updating. This partial coverage means that even when the central logic is correctly refactored, the test suite fails because downstream dependencies remain inconsistent.

The right panel separates resolved (solid) from unresolved (dashed) instances. For both models, successful resolutions complete in markedly fewer interaction rounds: the pass curves rise steeply and plateau early, while the fail curves are shifted rightward and more gradual. This gap reveals that agents entering unproductive cycles---repeatedly reading files, attempting edits, encountering test failures, and reverting---consume many rounds without expanding the scope of their modifications. In contrast, successful agents exhibit focused behavior, efficiently identifying and modifying the full set of required files in fewer steps. Together with the file-count analysis, these patterns suggest that the key bottleneck is not reasoning ability per se, but rather the capacity to maintain a coherent plan across many files and persist through the cascading consequences of a large-scale structural change.

\subsection{Cost and efficiency analysis}

\textcolor{black}{Table~\ref{tab:main-results} reports the average number of agent steps and API cost per instance for each model. A striking pattern emerges: \emph{higher cost does not translate to proportionally higher resolve rate}. Under OpenHands, Claude Sonnet 4.6 is the most expensive model (\$4.77 per instance, 117.9 steps on average) yet trails GPT-5.2 (41.2\% at \$3.60), and Gemini-3-Pro spends \$1.49 per instance for only 19.4\%. Open-weight models are dramatically more cost-efficient: GLM-5 resolves 36.5\% at just \$0.24 per instance---roughly one-twentieth the cost of Claude Sonnet 4.6---and Kimi-K2.5 achieves 32.9\% at \$0.72.}

\textcolor{black}{We also observe that a larger step budget does not guarantee progress. Manual inspection of trajectories shows that models facing large multi-file patches can enter repetitive edit--revert cycles---which we term \emph{unproductive exploration}---consuming many steps without expanding the scope of their modifications. Qwen3.5 represents an extreme case: it takes the most steps under both scaffolds (155.4 and 141.2 on average) yet never leads overall, and under mini-swe-agent it achieves the lowest resolve rate (20.6\%), suggesting that excessive exploration without effective cross-file coordination is counterproductive.}


\section{Conclusion}
\label{sec:conclusion}

We presented \approach, an expert-curated, multilingual code refactoring benchmark designed to address three critical gaps in current AI coding evaluation: the saturation of existing benchmarks, the lack of evaluation quality assurance, and the absence of refactoring as a benchmark domain. \approach comprises 170 instances drawn from real commits across seven programming languages (Python, Java, TypeScript, Go, C, C++, and Rust) and 70 repositories, selected from 29,782 initial candidates through a rigorous three-stage pipeline. Every instance undergoes expert curation: issue descriptions are rewritten from scratch to eliminate ambiguity and answer leakage, and test suites are manually reviewed to remove overly narrow tests that reject valid solutions and overly broad tests that check unstated requirements. 
Evaluation of six frontier models reveals that \approach remains far from saturated: \textcolor{black}{the best model (GPT-5.2) achieves only 41.2\% resolve rate}, and no model dominates across all languages. Our analysis uncovers two key findings. \textcolor{black}{First, open-weight models are highly competitive with proprietary ones on large-scale refactoring: GLM-5 and Qwen3.5 come within five points of the best proprietary model at a fraction of the cost, and manual trajectory inspection shows that higher spending often reflects unproductive exploration cycles rather than genuine progress.} Second, the dominant failure mode is incomplete refactoring: agents consistently modify fewer files than the gold patch requires, indicating that sustained cross-file coordination remains a fundamental bottleneck for current AI coding agents.


\subsubsection*{Acknowledgments}
%


This research was supported by the National Key Research and Development Program of China (Grant No. 2023YFB4503802), the Natural Science Foundation of Shanghai (Grant No. 25ZR1401175), the Hong Kong Research Grant Council General Research Fund (Grant No. 16206524), and the Hong Kong Research Grant Council Theme-based Research Scheme (Grant No. T41-517\_25-N). We thank Kexin Pei for valuable feedback and advice.

\section*{Ethics statement}

All repositories included in our benchmark are publicly available on GitHub and use approved open-source licenses (e.g., MIT, Apache 2.0, BSD), and we respect each project's licensing terms in our distribution. No personal or sensitive data is collected; our dataset consists solely of code, commit metadata, and test cases derived from public repositories. The benchmark is intended exclusively for the evaluation and improvement of AI coding agents, and we encourage responsible use that aligns with the goals of advancing software engineering research.

\section*{LLM usage disclosure}

In accordance with the policy on LLM usage, we disclose the following uses of large language models in this work: (1)~Human annotators collaboratively used LLMs as writing assistants to rewrite problem statements and review test suites during benchmark curation (Stage~3 of our pipeline); the LLM served as an interactive tool under human direction, not as an autonomous generator. (2)~Claude Sonnet 4.6 was used for multi-label classification of task categories and required reasoning skills (Appendix~\ref{app:categories}); these classifications are used for analysis only and do not affect the benchmark instances or evaluation results. (3)~LLMs were used for minor assistance in drafting and editing portions of this paper.

\bibliography{camera_ready}

@article{humaneval,
  title={Evaluating large language models trained on code},
  author={Chen, Mark and Tworek, Jerry and Jun, Heewoo and Yuan, Qiming and Pinto, Henrique Ponde De Oliveira and Kaplan, Jared and Edwards, Harri and Burda, Yuri and Joseph, Nicholas and Brockman, Greg and others},
  journal={arXiv preprint arXiv:2107.03374},
  year={2021}
}

@article{mbpp,
  title={Program synthesis with large language models},
  author={Austin, Jacob and Odena, Augustus and Nye, Maxwell and Bosma, Maarten and Michalewski, Henryk and Dohan, David and Jiang, Ellen and Cai, Carrie and Terry, Michael and Le, Quoc and others},
  journal={arXiv preprint arXiv:2108.07732},
  year={2021}
}

@article{livecodebench,
  title={Livecodebench: Holistic and contamination free evaluation of large language models for code},
  author={Jain, Naman and Han, King and Gu, Alex and Li, Wen-Ding and Yan, Fanjia and Zhang, Tianjun and Wang, Sida and Solar-Lezama, Armando and Sen, Koushik and Stoica, Ion},
  journal={arXiv preprint arXiv:2403.07974},
  year={2024}
}

@inproceedings{swe-bench,
  title={Swe-bench: Can language models resolve real-world github issues?},
  author={Jimenez, Carlos E and Yang, John and Wettig, Alexander and Yao, Shunyu and Pei, Kexin and Press, Ofir and Narasimhan, Karthik},
  booktitle={International Conference on Learning Representations},
  volume={2024},
  pages={54107--54157},
  year={2024}
}

@article{swe-bench-pro,
  title={Swe-bench pro: Can ai agents solve long-horizon software engineering tasks?},
  author={Deng, Xiang and Da, Jeff and Pan, Edwin and He, Yannis Yiming and Ide, Charles and Garg, Kanak and Lauffer, Niklas and Park, Andrew and Pasari, Nitin and Rane, Chetan and others},
  journal={arXiv preprint arXiv:2509.16941},
  year={2025}
}

@article{swe-evo,
  title={SWE-EVO: Benchmarking coding agents in long-horizon software evolution scenarios},
  author={Le, Tue and Thai, Minh VT and Manh, Dung Nguyen and Nhat, Huy Phan and Bui, Nghi DQ},
  journal={arXiv preprint arXiv:2512.18470},
  year={2025}
}

@article{swe-rebench,
  title={Swe-rebench: An automated pipeline for task collection and decontaminated evaluation of software engineering agents},
  author={Badertdinov, Ibragim and Golubev, Alexander and Nekrashevich, Maksim and Shevtsov, Anton and Karasik, Simon and Andriushchenko, Andrei and Trofimova, Maria and Litvintseva, Daria and Yangel, Boris},
  journal={arXiv preprint arXiv:2505.20411},
  year={2025}
}

@article{swe-bench-live,
  title={Swe-bench goes live!},
  author={Zhang, Linghao and He, Shilin and Zhang, Chaoyun and Kang, Yu and Li, Bowen and Xie, Chengxing and Wang, Junhao and Wang, Maoquan and Huang, Yufan and Fu, Shengyu and others},
  journal={Advances in Neural Information Processing Systems},
  volume={38},
  year={2026}
}

@article{multi-swe-bench,
  title={Multi-swe-bench: A multilingual benchmark for issue resolving},
  author={Zan, Daoguang and Huang, Zhirong and Liu, Wei and Chen, Hanwu and Xin, Shulin and Zhang, Linhao and Liu, Qi and Aoyan, Li and Chen, Lu and Zhong, Xiaojian and others},
  journal={Advances in Neural Information Processing Systems},
  volume={38},
  year={2026}
}

@article{swe-factory,
  title={Swe-factory: Your automated factory for issue resolution training data and evaluation benchmarks},
  author={Guo, Lianghong and Wang, Yanlin and Li, Caihua and Tao, Wei and Yang, Pengyu and Chen, Jiachi and Song, Haoyu and Tang, Duyu and Zheng, Zibin},
  journal={arXiv preprint arXiv:2506.10954},
  year={2025}
}

@article{terminal-bench,
  title={Terminal-bench: Benchmarking agents on hard, realistic tasks in command line interfaces},
  author={Merrill, Mike A and Shaw, Alexander G and Carlini, Nicholas and Li, Boxuan and Raj, Harsh and Bercovich, Ivan and Shi, Lin and Shin, Jeong Yeon and Walshe, Thomas and Buchanan, E Kelly and others},
  journal={arXiv preprint arXiv:2601.11868},
  year={2026}
}

@article{refactorbench,
  title={Refactorbench: Evaluating stateful reasoning in language agents through code},
  author={Gautam, Dhruv and Garg, Spandan and Jang, Jinu and Sundaresan, Neel and Moghaddam, Roshanak Zilouchian},
  journal={arXiv preprint arXiv:2503.07832},
  year={2025}
}

@article{swe-refactor,
  title={SWE-Refactor: A Repository-Level Benchmark for Real-World LLM-Based Code Refactoring},
  author={Xu, Yisen and Yang, Jinqiu and others},
  journal={arXiv preprint arXiv:2602.03712},
  year={2026}
}

@article{agentic-refactoring,
  title={Agentic Refactoring: An Empirical Study of AI Coding Agents},
  author={Horikawa, Kosei and Li, Hao and Kashiwa, Yutaro and Adams, Bram and Iida, Hajimu and Hassan, Ahmed E},
  journal={arXiv preprint arXiv:2511.04824},
  year={2025}
}

@article{automated-refactoring-discovery,
  title={An automated approach to discovering software refactorings by comparing successive versions},
  author={Liu, Bo and Liu, Hui and Niu, Nan and Zhang, Yuxia and Li, Guangjie and Jiang, He and Jiang, Yanjie},
  journal={IEEE Transactions on Software Engineering},
  volume={51},
  number={5},
  pages={1358--1380},
  year={2025},
  publisher={IEEE}
}

@article{empirical-refactoring-study,
  title={An Empirical Study of Software Refactorings in Real-World Open-Source Java Projects},
  author={Nyirongo, Bridget and Jiang, Yanjie and Niu, Nan and Liu, Hui},
  journal={IEEE Transactions on Software Engineering},
  year={2025},
  publisher={IEEE}
}

@misc{gpt5-codex,
  title     = {{GPT-5.1-Codex-Max}},
  author    = {{OpenAI}},
  year      = {2025},
  howpublished = {\url{https://openai.com/index/gpt-5-1-codex-max/}},
  note      = {Accessed: 2025-11-19}
}

@misc{gpt52-codex,
  title     = {Introducing Upgrades to {Codex}},
  author    = {{OpenAI}},
  year      = {2025},
  howpublished = {\url{https://openai.com/index/introducing-upgrades-to-codex/}},
  note      = {Accessed: 2025-09-15}
}

@misc{openai-swebench-deprecated,
  title     = {Why {SWE}-bench {Verified} No Longer Measures Frontier Coding Capabilities},
  author    = {{OpenAI}},
  year      = {2026},
  howpublished = {\url{https://openai.com/index/why-we-no-longer-evaluate-swe-bench-verified/}},
  note      = {Accessed: 2026-02-23}
}

@misc{cursorbench,
  title     = {How We Compare Model Quality at {Cursor}},
  author    = {Naman Jain},
  year      = {2026},
  howpublished = {\url{https://cursor.com/blog/cursorbench}},
  note      = {Accessed: 2026-03-11}
}

@article{metr-time-horizons,
  title={Measuring ai ability to complete long tasks},
  author={Kwa, Thomas and West, Ben and Becker, Joel and Deng, Amy and Garcia, Katharyn and Hasin, Max and Jawhar, Sami and Kinniment, Megan and Rush, Nate and Von Arx, Sydney and others},
  journal={arXiv preprint arXiv:2503.14499},
  volume={352},
  year={2025},
  publisher={Mar}
}

@inproceedings{longcli-bench,
  title={Longcli-bench: A preliminary benchmark and study for long-horizon agentic programming in command-line interfaces},
  author={Feng, Yukang and Sun, Jianwen and Yang, Zelai and Ai, Jiaxin and Li, Chuanhao and Li, Zizhen and Zhang, Fanrui and He, Kang and Ma, Rui and Lin, Jifan and others},
  booktitle={Findings of the Association for Computational Linguistics: ACL 2026},
  pages={29952--29963},
  year={2026}
}

@article{qwen3-coder,
  title={Qwen3-coder-next technical report},
  author={Cao, Ruisheng and Chen, Mouxiang and Chen, Jiawei and Cui, Zeyu and Feng, Yunlong and Hui, Binyuan and Jing, Yuheng and Li, Kaixin and Li, Mingze and Lin, Junyang and others},
  journal={arXiv preprint arXiv:2603.00729},
  year={2026}
}

@article{deepseek-coder-v2,
  title={Deepseek-coder-v2: Breaking the barrier of closed-source models in code intelligence},
  author={Zhu, Qihao and Guo, Daya and Shao, Zhihong and Yang, Dejian and Wang, Peiyi and Xu, Runxin and Wu, Y and Li, Yukun and Gao, Huazuo and Ma, Shirong and others},
  journal={arXiv preprint arXiv:2406.11931},
  year={2024}
}

@article{glm5,
  title={Glm-5: from vibe coding to agentic engineering},
  author={Zeng, Aohan and Lv, Xin and Hou, Zhenyu and Du, Zhengxiao and Zheng, Qinkai and Chen, Bin and Yin, Da and Ge, Chendi and Huang, Chenghua and Xie, Chengxing and others},
  journal={arXiv preprint arXiv:2602.15763},
  year={2026}
}

@article{kimi-k2,
  title={Kimi k2: Open agentic intelligence},
  author={Team, Kimi and Bai, Yifan and Bao, Yiping and Charles, Y and Chen, Cheng and Chen, Guanduo and Chen, Haiting and Chen, Huarong and Chen, Jiahao and Chen, Ningxin and others},
  journal={arXiv preprint arXiv:2507.20534},
  year={2025}
}

@article{gpt5,
  title={Openai gpt-5 system card},
  author={Singh, Aaditya and Fry, Adam and Perelman, Adam and Tart, Adam and Ganesh, Adi and El-Kishky, Ahmed and McLaughlin, Aidan and Low, Aiden and Ostrow, AJ and Ananthram, Akhila and others},
  journal={arXiv preprint arXiv:2601.03267},
  year={2025}
}

@misc{gpt52,
  title        = {{GPT-5} System Card Update: {GPT-5.2}},
  author       = {{OpenAI}},
  year         = {2025},
  howpublished = {\url{https://openai.com/index/gpt-5-system-card-update-gpt-5-2/}}
}

@misc{gemini3pro,
  title        = {Gemini 3 Pro},
  author       = {{Google DeepMind}},
  year         = {2025},
  howpublished = {\url{https://blog.google/products-and-platforms/products/gemini/gemini-3/}}
}

@misc{claudesonnet46,
  title        = {Claude Sonnet 4.6},
  author       = {{Anthropic}},
  year         = {2026},
  howpublished = {\url{https://www.anthropic.com/news/claude-sonnet-4-6}}
}

@article{kimi-k25,
  title={Kimi K2. 5: Visual Agentic Intelligence},
  author={Team, Kimi and Bai, Tongtong and Bai, Yifan and Bao, Yiping and Cai, SH and Cao, Yuan and Charles, Y and Che, HS and Chen, Cheng and Chen, Guanduo and others},
  journal={arXiv preprint arXiv:2602.02276},
  year={2026}
}

@inproceedings{swe-agent,
  title={Swe-agent: Agent-computer interfaces enable automated software engineering},
  author={Yang, John and Jimenez, Carlos E and Wettig, Alexander and Lieret, Kilian and Yao, Shunyu and Narasimhan, Karthik R and Press, Ofir},
  booktitle={The Thirty-eighth Annual Conference on Neural Information Processing Systems},
  year={2024}
}

@inproceedings{openhands,
  title={Openhands: An open platform for ai software developers as generalist agents},
  author={Wang, Xingyao and Li, Boxuan and Song, Yufan and Xu, Frank F and Tang, Xiangru and Zhuge, Mingchen and Pan, Jiayi and Song, Yueqi and Li, Bowen and Singh, Jaskirat and others},
  booktitle={International Conference on Learning Representations},
  volume={2025},
  pages={65882--65919},
  year={2025}
}

@article{agentless,
  title={Agentless: Demystifying llm-based software engineering agents},
  author={Xia, Chunqiu Steven and Deng, Yinlin and Dunn, Soren and Zhang, Lingming},
  journal={arXiv preprint arXiv:2407.01489},
  year={2024}
}

@inproceedings{codetaste,
  title={CodeTaste: Can LLMs Generate Human-Level Code Refactorings?},
  author={Thillen, Alex and M{\"u}ndler, Niels and Raychev, Veselin and Vechev, Martin},
  booktitle={Forty-third International Conference on Machine Learning},
  year={2026}
}

@article{refactoring-llm-empirical,
  title={An empirical study on the code refactoring capability of large language models},
  author={Cordeiro, Jonathan and Noei, Shayan and Zou, Ying},
  journal={ACM Transactions on Software Engineering and Methodology},
  year={2024},
  publisher={ACM New York, NY}
}

@article{refactoring-llm-fewshot,
  title={Code Refactoring with LLM: A Comprehensive Evaluation With Few-Shot Settings},
  author={Tapader, Md Raihan and Rahman, Md Mostafizer and Shiplu, Ariful Islam and Amin, Md Faizul Ibne and Watanobe, Yutaka},
  journal={arXiv preprint arXiv:2511.21788},
  year={2025}
}

@article{refactoringminer,
  title={RefactoringMiner 2.0},
  author={Tsantalis, Nikolaos and Ketkar, Ameya and Dig, Danny},
  journal={IEEE Transactions on Software Engineering},
  volume={48},
  number={3},
  pages={930--950},
  year={2020},
  publisher={IEEE}
}

@article{swe-polybench,
  title={Swe-polybench: A multi-language benchmark for repository level evaluation of coding agents},
  author={Rashid, Muhammad Shihab and Bock, Christian and Zhuang, Yuan and Buchholz, Alexander and Esler, Tim and Valentin, Simon and Franceschi, Luca and Wistuba, Martin and Sivaprasad, Prabhu Teja and Kim, Woo Jung and others},
  journal={arXiv preprint arXiv:2504.08703},
  year={2025}
}

@article{zeng2025pruning,
  title={Pruning the unsurprising: Efficient code reasoning via first-token surprisal},
  author={Zeng, Wenhao and Wang, Yaoning and Hu, Chao and Shi, Yuling and Wan, Chengcheng and Zhang, Hongyu and Gu, Xiaodong},
  journal={arXiv preprint arXiv:2508.05988},
  year={2025}
}

@article{hu2026line,
  title={In line with context: Repository-level code generation via context inlining},
  author={Hu, Chao and Zeng, Wenhao and Shi, Yuling and Shen, Beijun and Gu, Xiaodong},
  journal={Proceedings of the ACM on Software Engineering},
  volume={3},
  number={FSE},
  pages={1469--1491},
  year={2026},
  publisher={ACM New York, NY, USA}
}

@article{zeng2026glimprouter,
  title={Glimprouter: Efficient collaborative inference by glimpsing one token of thoughts},
  author={Zeng, Wenhao and Zhang, Xuteng and Shi, Yuling and Hu, Chao and Chen, Yuting and Shen, Beijun and Gu, Xiaodong},
  journal={arXiv preprint arXiv:2601.05110},
  year={2026}
}

@article{zeng2026dockerless,
  title={Dockerless: Environment-Free Program Verifier for Coding Agents},
  author={Zeng, Wenhao and Shi, Yuling and Gu, Xiaodong and Hu, Chao and Wang, Chaofan and Cui, Yuhao and Zhou, Hongting and Qi, Mengnan and Wangni, Jianqiao and Yu, Zhaojian and others},
  journal={arXiv preprint arXiv:2606.28436},
  year={2026}
}

@article{gao2026swe,
  title={SWE-MeM: Learning Adaptive Memory Management for Long-Horizon Coding Agents},
  author={Gao, Shuzheng and Zeng, Wenhao and Yu, Zhaojian and Wangni, Jianqiao and Wang, Chaozheng and Cai, Kai and He, Shilin and Lyu, Michael R},
  journal={arXiv preprint arXiv:2606.28434},
  year={2026}
}

@article{huang2026deepswe,
  title={DeepSWE: Measuring Frontier Coding Agents on Original, Long-Horizon Engineering Tasks},
  author={Huang, Wenqi and Lee, Charley and Tng, Leonard and Ge, Serena},
  journal={arXiv preprint arXiv:2607.07946},
  year={2026}
}

@article{zhao2026immersion,
  title={Immersion in the github universe: Scaling coding agents to mastery},
  author={Zhao, Jiale and Chen, Guoxin and Meng, Fanzhe and Li, Minghao and Chen, Jie and Xu, Hui and Sun, Yongshuai and Zhao, Wayne Xin and Song, Ruihua and Zhang, Yuan and others},
  journal={arXiv preprint arXiv:2602.09892},
  year={2026}
}

@article{fu2026davinci,
  title={davinci-env: Open swe environment synthesis at scale},
  author={Fu, Dayuan and Wu, Shenyu and Wu, Yunze and Peng, Zerui and Huang, Yaxing and Sun, Jie and Zeng, Ji and Jiang, Mohan and Zhang, Lin and Li, Yukun and others},
  journal={arXiv preprint arXiv:2603.13023},
  year={2026}
}

@article{desai2026swe,
  title={SWE-Marathon: Can Agents Autonomously Complete Ultra-Long-Horizon Software Work?},
  author={Desai, Rishi and Hu, Jesse and Cabezas, Joan and Harsola, Neel and Shukla, Pratyush and Chaim, Roey Ben and Assadi, Adnan El and Kamath, Omkaar Mukund and Faldu, Fenil and Hebbar, Prannay and others},
  journal={arXiv preprint arXiv:2606.07682},
  year={2026}
}

@inproceedings{shi2024between,
  title={Between lines of code: Unraveling the distinct patterns of machine and human programmers},
  author={Shi, Yuling and Zhang, Hongyu and Wan, Chengcheng and Gu, Xiaodong},
  booktitle={2025 IEEE/ACM 47th International Conference on Software Engineering (ICSE 2025)},
  year={2024}
}

@inproceedings{shi2024code,
  title={From code to correctness: Closing the last mile of code generation with hierarchical debugging},
  author={Shi, Yuling and Wang, Songsong and Wan, Chengcheng and Wang, Min and Gu, Xiaodong},
  booktitle={2026 IEEE/ACM 48th International Conference on Software Engineering (ICSE 2026)},
  year={2024}
}

@article{peng2025swe,
  title={SWE-QA: Can Language Models Answer Repository-level Code Questions?},
  author={Peng, Weihan and Shi, Yuling and Wang, Yuhang and Zhang, Xinyun and Shen, Beijun and Gu, Xiaodong},
  journal={arXiv preprint arXiv:2509.14635},
  year={2025}
}

@inproceedings{shi2025longcodezip,
  title={LongCodeZip: Compress Long Context for Code Language Models},
  author={Shi, Yuling and Qian, Yichun and Zhang, Hongyu and Shen, Beijun and Gu, Xiaodong},
  booktitle={2025 IEEE/ACM 40th International Conference on Automated Software Engineering (ASE)},
  year={2025}
}

@article{shi2026codeocr,
  title={CodeOCR: On the Effectiveness of Vision Language Models in Code Understanding},
  author={Shi, Yuling and Xie, Chaoxiang and Sun, Zhensu and Chen, Yeheng and Zhang, Chenxu and Yun, Longfei and Wan, Chengcheng and Zhang, Hongyu and Lo, David and Gu, Xiaodong},
  journal={arXiv preprint arXiv:2602.01785},
  year={2026}
}

@article{chen2025swe,
  title={SWE-Exp: Experience-driven software issue resolution},
  author={Chen, Silin and Lin, Shaoxin and Shi, Yuling and Lian, Heng and Gu, Xiaodong and Yun, Longfei and Chen, Dong and Cao, Lin and Liu, Jiyang and Xia, Nu and others},
  journal={arXiv preprint arXiv:2507.23361},
  year={2025}
}

@article{wang2026swe,
  title={Swe-pruner: Self-adaptive context pruning for coding agents},
  author={Wang, Yuhang and Shi, Yuling and Yang, Mo and Zhang, Rongrui and He, Shilin and Lian, Heng and Chen, Yuting and Ye, Siyu and Cai, Kai and Gu, Xiaodong},
  journal={arXiv preprint arXiv:2601.16746},
  year={2026}
}

@article{li2025swe,
  title={Swe-debate: Competitive multi-agent debate for software issue resolution},
  author={Li, Han and Shi, Yuling and Lin, Shaoxin and Gu, Xiaodong and Lian, Heng and Wang, Xin and Jia, Yantao and Huang, Tao and Wang, Qianxiang},
  journal={arXiv preprint arXiv:2507.23348},
  year={2025}
}

@article{zhang2026fastcontext,
  title={FastContext: Training Efficient Repository Explorer for Coding Agents},
  author={Zhang, Shaoqiu and Wang, Maoquan and Shi, Yuling and Wang, Yuhang and Gu, Xiaodong and Yao, Yongqiang and Gong, Tori and Chen, Sheng and Fu, Rao and Agarwal, Anisha and others},
  journal={arXiv preprint arXiv:2606.14066},
  year={2026}
}

@article{shi2026earlyeval,
  title={EarlyEval: Cheaper Agent Evaluation via Early Outcome Prediction},
  author={Shi, Yuling and Sun, Zhensu and Dong, Junsen and Wan, Chengcheng and Lo, David and Gu, Xiaodong},
  journal={arXiv preprint arXiv:2609.02783},
  year={2026}
}

@misc{shi2026speculativeforkingturningidle,
      title={Speculative Forking: Turning Idle Wait into Parallel Exploration for Repository Environment Setup}, 
      author={Yuling Shi and Mingchu Zhong and Zhensu Sun and Weihan Peng and Zhi Chen and Shilin He and Chengcheng Wan and Kai Cai and Hongyu Zhang and David Lo and Xiaodong Gu},
      year={2026},
      eprint={2610.33129},
      archivePrefix={arXiv},
      primaryClass={cs.SE},
      url={https://arxiv.org/abs/2610.33129}, 
}
\bibliographystyle{plainnat}

\appendix

\newpage
\section{Additional dataset details}
\label{app:data}

\small
\begin{longtable}{lll}
\caption{Repositories and licenses used in \approach, grouped by language. 70 repositories across 7 languages.}
\label{tab:grouped-repos} \\
\toprule
\textbf{Language} & \textbf{Repository} & \textbf{License} \\
\midrule
\endfirsthead
\toprule
\textbf{Language} & \textbf{Repository} & \textbf{License} \\
\midrule
\endhead
\midrule
\multicolumn{3}{r}{\textit{Continued on next page}} \\
\endfoot
\bottomrule
\endlastfoot
C & betaflight/betaflight & GPL-3.0 \\
 & aviggiano/redis-roaring & MIT \\
 & davidesantangelo/krep & BSD-2-Clause \\
 & radareorg/radare2 & LGPL-3.0 \\
 & CESNET/libyang & BSD-3-Clause \\
 & arkq/bluez-alsa & MIT \\
 & aws/s2n-tls & Apache-2.0 \\
 & bitcoin-core/secp256k1 & MIT \\
 & openssl/openssl & Apache-2.0 \\
\midrule
C++ & deskflow/deskflow & GPL-2.0 \\
 & ETLCPP/etl & MIT \\
 & nasa/fprime & Apache-2.0 \\
 & Icinga/icinga2 & GPL-3.0 \\
 & LMMS/lmms & GPL-2.0 \\
 & OpenOrienteering/mapper & GPL-3.0 \\
 & WasmEdge/WasmEdge & Apache-2.0 \\
 & bloomberg/blazingmq & Apache-2.0 \\
 & biojppm/rapidyaml & MIT \\
\midrule
Go & cli/cli & MIT \\
 & go-gitea/gitea & MIT \\
 & TecharoHQ/anubis & MIT \\
 & restic/restic & BSD-2-Clause \\
 & OpenListTeam/OpenList & AGPL-3.0 \\
 & caddyserver/caddy & Apache-2.0 \\
 & derailed/k9s & Apache-2.0 \\
 & gitleaks/gitleaks & MIT \\
 & gohugoio/hugo & Apache-2.0 \\
 & grpc/grpc-go & Apache-2.0 \\
 & istio/istio & Apache-2.0 \\
 & jesseduffield/lazygit & MIT \\
 & kubernetes/kubernetes & Apache-2.0 \\
 & rqlite/rqlite & MIT \\
 & samber/lo & MIT \\
 & trufflesecurity/trufflehog & AGPL-3.0 \\
\midrule
Java & bazelbuild/bazel & Apache-2.0 \\
 & plantuml/plantuml & GPL-3.0 \\
 & hibernate/hibernate-orm & Apache-2.0 \\
 & apache/hbase & Apache-2.0 \\
 & apache/fesod & Apache-2.0 \\
 & apache/iceberg & Apache-2.0 \\
 & apache/maven & Apache-2.0 \\
 & apache/pinot & Apache-2.0 \\
 & google/gson & Apache-2.0 \\
 & alibaba/nacos & Apache-2.0 \\
 & swagger-api/swagger-core & Apache-2.0 \\
\midrule
Python & confident-ai/deepeval & Apache-2.0 \\
 & google/adk-python & Apache-2.0 \\
 & optuna/optuna & MIT \\
 & stanfordnlp/dspy & MIT \\
 & vibrantlabsai/ragas & Apache-2.0 \\
 & albumentations-team/albumentations & MIT \\
 & huggingface/transformers & Apache-2.0 \\
 & langchain-ai/langchain & MIT \\
 & verl-project/verl & Apache-2.0 \\
 & django/django & BSD-3-Clause \\
 & google/langextract & Apache-2.0 \\
 & huggingface/lerobot & Apache-2.0 \\
 & hummingbot/hummingbot & Apache-2.0 \\
 & icloud-photos-downloader/icloud\_photos\_downloader & MIT \\
 & mikf/gallery-dl & GPL-2.0 \\
 & pandas-dev/pandas & BSD-3-Clause \\
 & pypa/pipenv & MIT \\
 & roboflow/supervision & MIT \\
\midrule
Rust & astral-sh/ruff & MIT \\
 & rust-lang/cargo & MIT/Apache-2.0 \\
 & openai/codex & Apache-2.0 \\
 & qdrant/qdrant & Apache-2.0 \\
 & tracel-ai/burn & MIT/Apache-2.0 \\
\midrule
TypeScript & angular/angular & MIT \\
 & ant-design/ant-design & MIT \\
\end{longtable}

The 70 repositories span a broad range of open-source licenses---predominantly Apache-2.0 and MIT, with representation from GPL, BSD, and AGPL---reflecting the diversity of the open-source ecosystem. Repository concentration varies by language: Go draws from 16 distinct repositories yielding 23 instances, while TypeScript's 28 instances come from only 2 repositories, with Angular alone contributing 25. This skew is a natural consequence of selecting large, actively maintained projects with substantial refactoring activity.

\subsection{Per-language summary}

\begin{table}[h]
    \centering
    \small
    \begin{tabular}{lccccc}
        \toprule
        \textbf{Language} & \textbf{\#Repos} & \textbf{\#Inst.} & \textbf{Avg. \#Files} & \textbf{Avg. LOC} & \textbf{Avg. \#Non-test} \\
        \midrule
        C & 9 & 20 & 17.9 & 424.1 & 15.2 \\
        C++ & 9 & 22 & 21.4 & 196.3 & 16.0 \\
        Go & 16 & 23 & 16.0 & 227.4 & 9.4 \\
        Java & 11 & 26 & 20.8 & 309.8 & 16.8 \\
        Python & 18 & 29 & 10.6 & 299.8 & 7.0 \\
        Rust & 5 & 22 & 14.5 & 284.8 & 11.0 \\
        TypeScript & 2 & 28 & 11.9 & 122.6 & 7.5 \\
        \midrule
        Overall & 70 & 170 & 15.9 & 261.6 & 11.5 \\
        \bottomrule
    \end{tabular}
    \caption{Per-language statistics for \approach, including number of repositories, instances, and average patch complexity.}
    \label{tab:app-language-summary}
\end{table}

\textcolor{black}{Java and C++ exhibit markedly larger patches, averaging 20.8 and 21.4 modified files per instance respectively, with C having the highest average LOC (424.1)---reflecting the cross-cutting nature of refactoring in codebases with deep type hierarchies and extensive header dependencies.} \textcolor{black}{In contrast, higher-level languages such as Python, Rust, and TypeScript average 10.6--14.5 files per instance, yet still require non-trivial multi-file coordination that distinguishes \approach from single-file benchmarks.}

\section{Task category analysis}
\label{app:categories}

Real-world refactoring commits rarely involve a single type of change. To characterize the diversity of skills required by \approach, we use Claude Sonnet 4.6 to perform multi-label classification of each instance into ten predefined categories: API Interface Change, Refactoring Cleanup, Bug Fix, New Feature, Documentation, Error Handling, Performance Optimization, Dependency Integration, Test Improvement, and Security Patch.

\begin{figure}[t]
    \centering
    \includegraphics[width=0.8\linewidth]{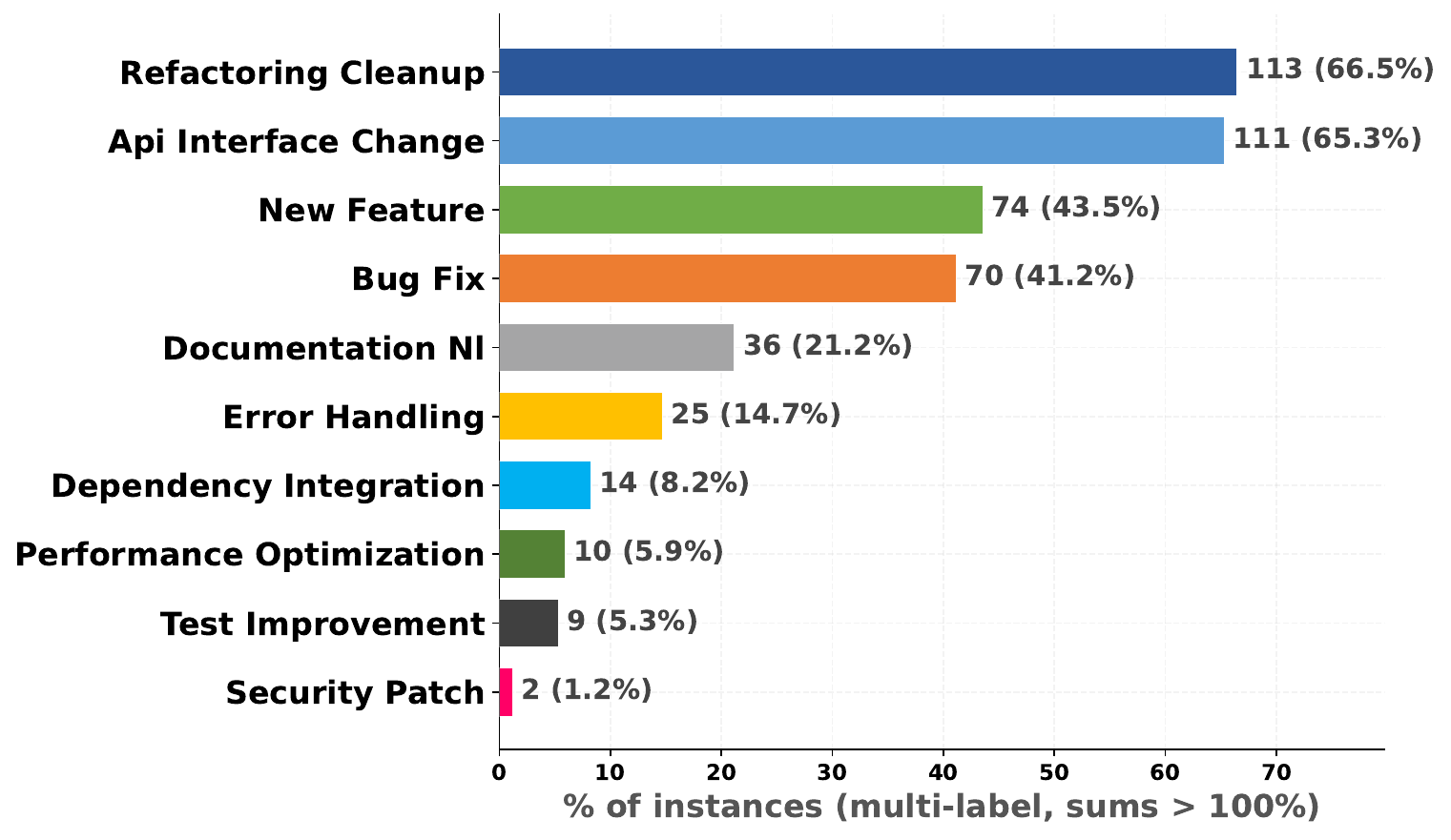}
    \caption{Distribution of task categories across \approach instances (multi-label; percentages sum to more than 100\%). The dominant categories---Refactoring Cleanup (66.5\%) and API Interface Change (65.3\%)---confirm the refactoring focus of the benchmark, while the substantial presence of New Feature (43.5\%) and Bug Fix (41.2\%) reflects the multi-faceted nature of real-world code restructuring.}
    \label{fig:category-distribution}
\end{figure}

\begin{figure}[h]
    \centering
    \includegraphics[width=0.65\linewidth]{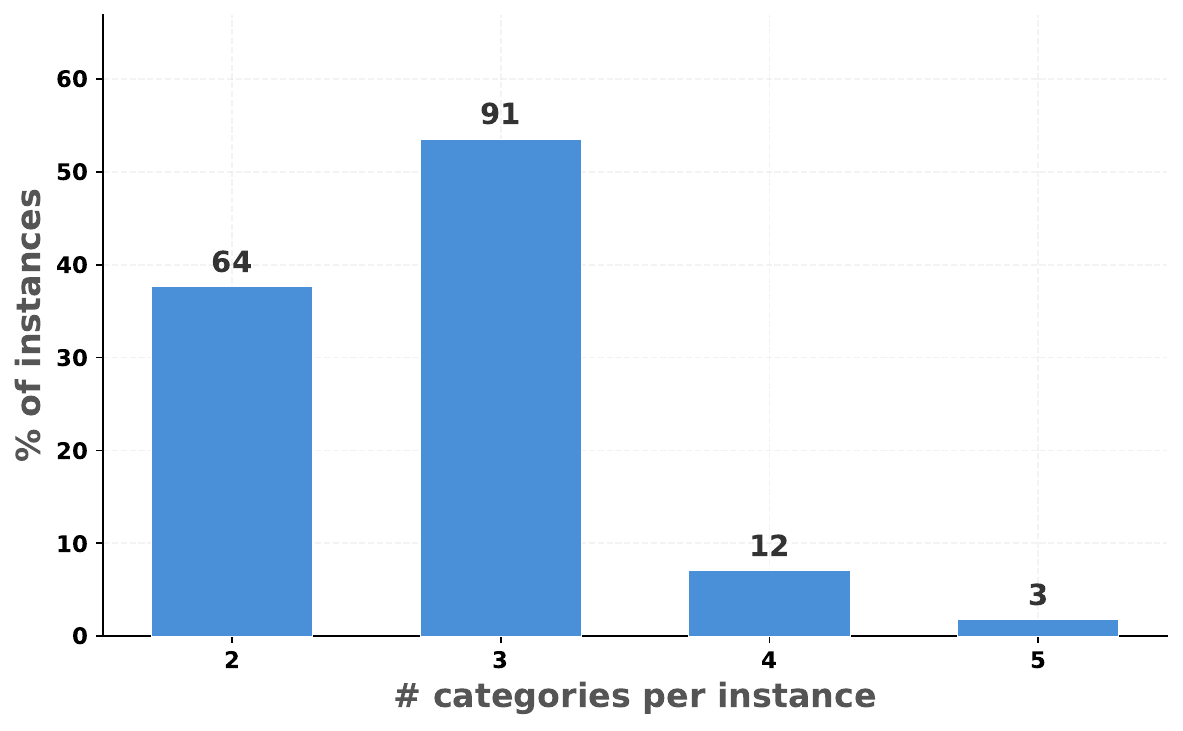}
    \caption{Number of categories per instance. \textcolor{black}{Every instance involves at least two categories, and nearly two-thirds (62.4\%) involve three or more simultaneously}, indicating that \approach tasks require holistic software engineering skills rather than isolated refactoring ability.}
    \label{fig:categories-per-instance}
\end{figure}

\begin{figure}[h]
    \centering
    \includegraphics[width=0.85\linewidth]{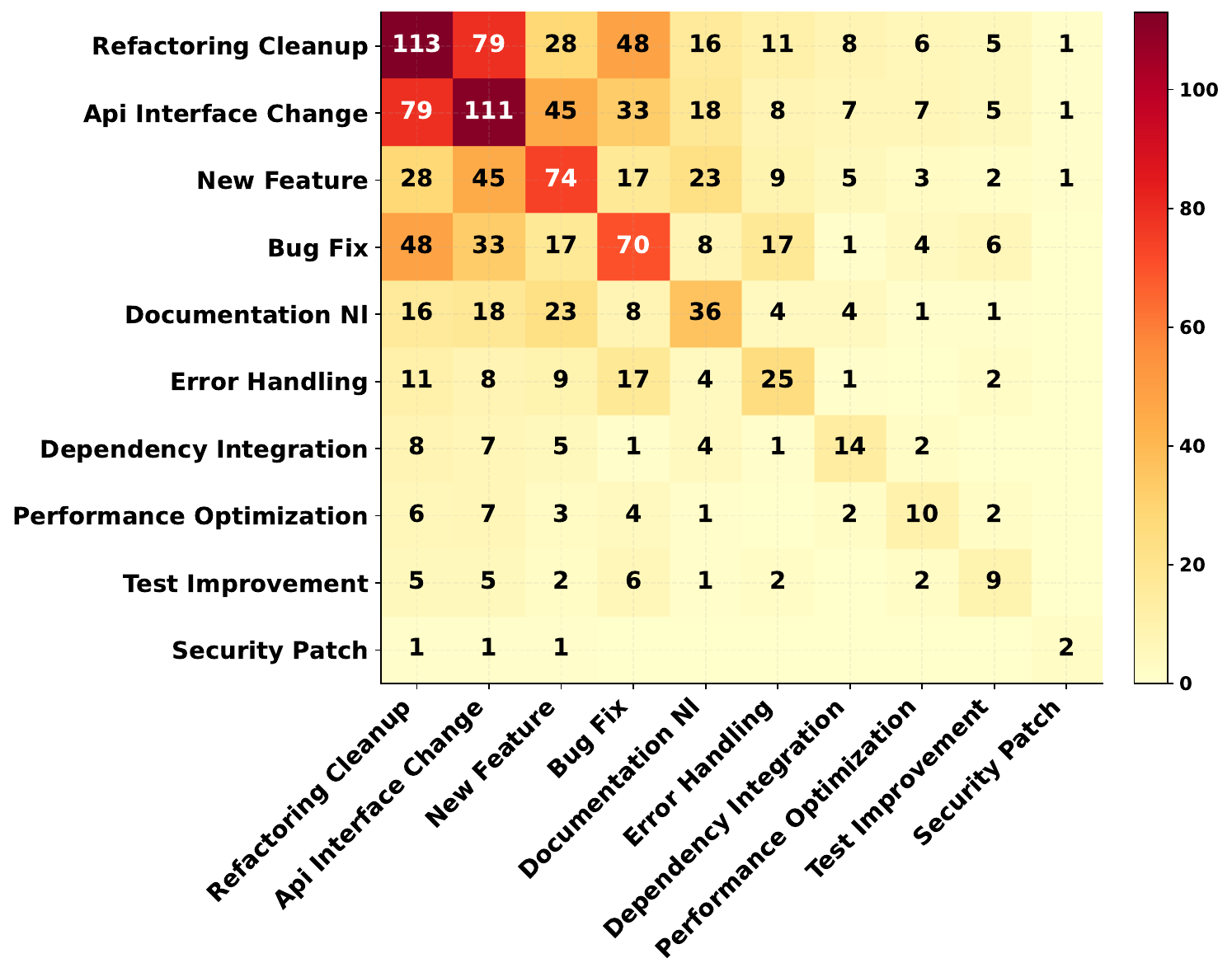}
    \caption{Co-occurrence matrix of task categories. API Interface Change and Refactoring Cleanup co-occur in 79 instances, while Bug Fix frequently accompanies both (33 and 48 instances respectively), reflecting how refactoring often surfaces latent defects that must be addressed concurrently.}
    \label{fig:category-cooccurrence}
\end{figure}

Figure~\ref{fig:category-distribution} shows that while refactoring-related categories dominate---Refactoring Cleanup appears in 66.5\% of instances and API Interface Change in 65.3\%---a substantial fraction of instances simultaneously involve new features (43.5\%), bug fixes (41.2\%), or documentation updates (21.2\%). As Figure~\ref{fig:categories-per-instance} illustrates, \textcolor{black}{62.4\% of instances span three or more categories}, with no instance involving fewer than two. The co-occurrence matrix (Figure~\ref{fig:category-cooccurrence}) reveals strong coupling between API changes and refactoring cleanup (79 co-occurrences), as well as between refactoring and bug fixing (48 co-occurrences)---patterns consistent with the empirical finding that restructuring code frequently exposes latent defects~\citep{empirical-refactoring-study}. This multi-faceted nature distinguishes \approach from benchmarks that test isolated skills and better reflects the compound challenges that developers face in practice.

The high prevalence of Bug Fix (\textcolor{black}{41.2\%}) is particularly noteworthy. Empirical studies of software maintenance have long observed that refactoring and bug fixing are deeply intertwined: restructuring code frequently exposes latent defects that were masked by the original design, and developers routinely address these defects within the same commit rather than deferring them to a separate change~\citep{empirical-refactoring-study}. This coupling means that an agent attempting to resolve a refactoring task in \approach must not only apply the intended structural transformation but also recognize and correctly fix any bugs that surface during the process~\citep{shi2024code}---a compound challenge that synthetic benchmarks, which typically isolate refactoring from bug fixing, cannot reproduce. Similarly, the presence of New Feature (\textcolor{black}{43.5\%}) reflects cases where refactoring serves as a prerequisite for introducing new capabilities: the structural improvement enables or unblocks a feature addition that is delivered in the same commit. These overlapping concerns make \approach tasks substantially harder than they would be if each category were tested in isolation.

\begin{figure}[t]
    \centering
    \includegraphics[width=\linewidth]{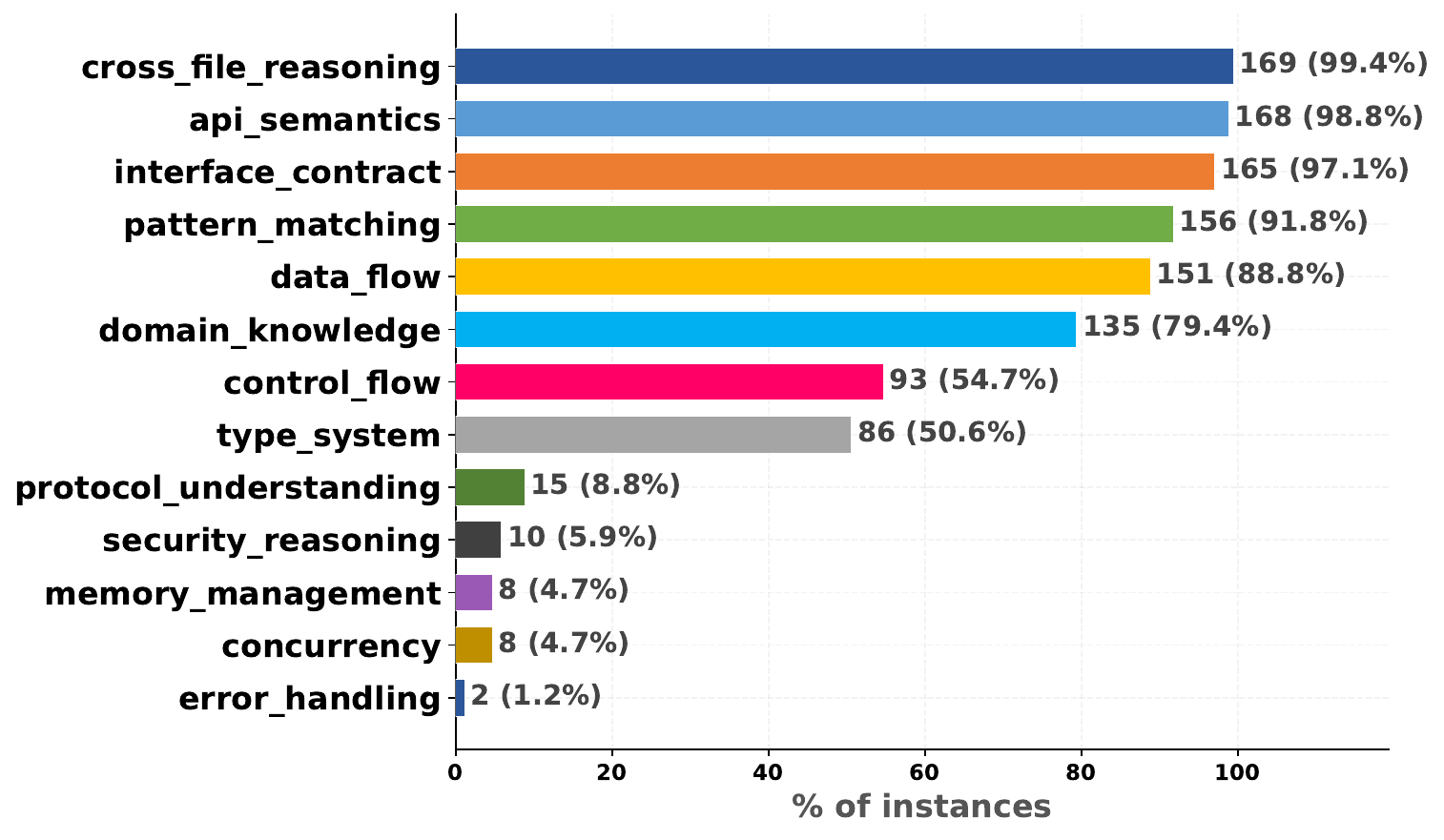}
    \caption{Distribution of required reasoning skills across \approach instances (multi-label). Nearly all instances require cross-file reasoning (99.4\%) and API semantics understanding (98.8\%), confirming that the benchmark systematically tests sustained multi-file comprehension. Pattern matching (91.8\%) and interface contract reasoning (97.1\%) are also near-universal, while data flow (88.8\%), domain knowledge (79.4\%) and type system reasoning (\textcolor{black}{50.6\%}) appear in the majority of instances.}
    \label{fig:required-skills}
\end{figure}

\paragraph{Required skills.}
Complementing the task-type analysis above, we also classify each instance by the reasoning skills required for successful resolution. Figure~\ref{fig:required-skills} shows the distribution.

The near-universal prevalence of cross-file reasoning (99.4\%) and API semantics (98.8\%) validates the benchmark's design goal of testing agents' ability to coordinate changes across file boundaries. The high frequency of pattern matching (91.8\%) reflects that refactoring tasks often require identifying and systematically transforming recurring code patterns throughout a codebase. Domain knowledge (79.4\%) and type system reasoning (50.6\%) are also common, indicating that many instances demand understanding of project-specific conventions or language-specific type constraints beyond generic code manipulation.

\section{Representative instances}
\label{app:cases}

We present one representative instance per language to illustrate the scale and diversity of \approach. For each instance we show the repository, number of modified files and lines of code, the commit URL for full inspection, an abbreviated file tree, and a summary of the rewritten problem statement. Table~\ref{tab:case-summary} provides an overview.

\begin{table}[t]
    \centering
    \small
    \resizebox{\textwidth}{!}{%
    \begin{tabular}{llrrl}
        \toprule
        \textbf{Lang.} & \textbf{Repository} & \textbf{Files} & \textbf{LOC} & \textbf{Refactoring summary} \\
        \midrule
        C++ & nasa/fprime & 244 & 559 & Unify header includes across framework \\
        Java & plantuml/plantuml & 94 & 1{,}629 & Add hour-level time resolution to Gantt engine \\
        C & betaflight/betaflight & 62 & 846 & Rename motor protocol configuration fields \\
        Rust & tracel-ai/burn & 49 & 1{,}084 & Unify scalar arguments across tensor operations \\
        Go & OpenListTeam/OpenList & 47 & 608 & Refactor upload stream buffering across drivers \\
        Python & google/langextract & 30 & 1{,}960 & Centralize provider output-format handling \\
        TS & ant-design/ant-design & 27 & 97 & Unify \texttt{destroyOnHidden} across components \\
        \bottomrule
    \end{tabular}}
    \caption{Summary of representative instances in \approach, one per language, sorted by number of modified files.}
    \label{tab:case-summary}
\end{table}

\subsection{C++: nasa/fprime (244 files, 559 LOC)}
\label{app:case-cpp}

\noindent\textbf{Commit:} \url{https://github.com/nasa/fprime/pull/3422}

\noindent\textbf{File tree (excerpt):}
{\small
\begin{Verbatim}[commandchars=\\\{\}]
nasa/fprime (244 files, \textcolor{checkgreen}{+591}/\textcolor{black}{-514})
|-- Autocoders/Python/src/.../component/cpp.tmpl
|-- Autocoders/Python/src/.../impl/cpp.tmpl
|-- config/FpConfig.fpp
|-- config/FpConfig.h
|-- Fw/FPrimeBasicTypes.h              \textcolor{checkgreen}{[NEW]}
|-- Fw/FPrimeBasicTypes.hpp            \textcolor{checkgreen}{[NEW]}
|-- cmake/platform/unix/Platform/...   \textcolor{checkgreen}{[NEW]}
|-- Fw/Types/BasicTypes.h
|-- Drv/BlockDriver/BlockDriverImpl.cpp
|-- Svc/ActiveLogger/ActiveLoggerImpl.cpp
\textcolor{gray}{|-- ... (234 more files)}
\end{Verbatim}
}

\noindent\textbf{Problem statement (first paragraph):}
\begin{quote}
\small\itshape
During routine development and maintenance of the F'Prime framework, teams have observed growing friction around header organization and build dependencies. The monolithic \texttt{FpConfig.hpp} header has accumulated responsibilities spanning basic fixed-width types (like \texttt{I32}, \texttt{U64}), platform-specific type configurations, project-level aliases, framework constants, and build-time switches. This conflation of concerns creates tangible workflow disruptions: any modification---even a minor adjustment to a single alias or configuration value---triggers near-total recompilation across the entire framework and dependent projects, significantly extending build cycles during iterative development.

[\ldots]
\end{quote}

\subsection{Java: plantuml/plantuml (94 files, 1{,}629 LOC)}
\label{app:case-java}

\noindent\textbf{Commit:} \url{https://github.com/plantuml/plantuml/commit/c910f8b}

\noindent\textbf{File tree (excerpt):}
{\small
\begin{Verbatim}[commandchars=\\\{\}]
plantuml/plantuml (94 files, \textcolor{checkgreen}{+1550}/\textcolor{black}{-2212})
|-- src/.../chronology/ChronologyDiagram.java
|-- src/.../chronology/ComplementHour.java
|-- src/.../chronology/HourPattern.java
|-- src/.../project/ConstantPlan.java
|-- src/.../project/GanttDiagram.java
|-- src/.../project/Load.java
|-- src/.../project/OpenClose.java
|-- src/.../project/time3/Day.java
\textcolor{gray}{|-- ... (86 more files)}
\end{Verbatim}
}

\noindent\textbf{Problem statement (first paragraph):}
\begin{quote}
\small\itshape
The Gantt diagram engine in PlantUML has historically operated exclusively at day-level resolution, treating each calendar day as an indivisible atomic unit for all scheduling calculations. While sufficient for high-level project visualization, this architectural constraint increasingly limits the system's ability to support evolving user requirements involving finer temporal precision. Users attempting to model scenarios such as tasks spanning partial days (e.g., ``9 AM to 3 PM''), resources with intra-day availability patterns, or dependencies requiring hour-level alignment encounter fundamental limitations. The current implementation forces approximations---like splitting single-day efforts across multiple artificial days---which introduce inaccuracies in duration calculations, resource load reporting, and constraint validation. These approximations become especially problematic when integrating with external tools that export time data with sub-day precision or when users require precise effort tracking across non-standard work intervals.

[\ldots]
\end{quote}

\subsection{C: betaflight/betaflight (62 files, 846 LOC)}
\label{app:case-c}

\noindent\textbf{Commit:} \url{https://github.com/betaflight/betaflight/commit/9230c5f}

\noindent\textbf{File tree (excerpt):}
{\small
\begin{Verbatim}[commandchars=\\\{\}]
betaflight/betaflight (62 files, \textcolor{checkgreen}{+971}/\textcolor{black}{-806})
|-- mk/source.mk
|-- src/main/blackbox/blackbox.c
|-- src/main/cli/cli.c
|-- src/main/config/config.c
|-- src/main/drivers/dshot.c
|-- src/main/drivers/motor.c
|-- src/main/drivers/motor.h
|-- src/main/drivers/motor_types.h          \textcolor{checkgreen}{[NEW]}
|-- src/main/drivers/pwm_output.c           \textcolor{checkgreen}{[NEW]}
|-- src/platform/common/stm32/...           \textcolor{checkgreen}{[NEW]}
\textcolor{gray}{|-- ... (52 more files)}
\end{Verbatim}
}

\noindent\textbf{Problem statement (first paragraph):}
\begin{quote}
\small\itshape
The motor configuration subsystem in the \texttt{motorDevConfig\_t} structure uses field names that inaccurately imply PWM-specific scope for settings that apply across all motor protocol families. Specifically, the field \texttt{motorPwmProtocol} stores the motor protocol type for all protocols---including digital ones like Dshot and ProShot---yet its name suggests it is limited to PWM. Similarly, \texttt{motorPwmInversion} controls signal inversion for any motor output regardless of protocol family, and \texttt{useUnsyncedPwm} governs whether motor updates run continuously or are synchronized to the PID loop, a concept that is not inherently PWM-specific. These misleading names create cognitive friction during development and code review, and they cause confusion when configuring digital protocols where ``PWM'' terminology does not apply.

[\ldots]
\end{quote}

\subsection{Rust: tracel-ai/burn (49 files, 1{,}084 LOC)}
\label{app:case-rust}

\noindent\textbf{Commit:} \url{https://github.com/tracel-ai/burn/pull/4337}

\noindent\textbf{File tree (excerpt):}
{\small
\begin{Verbatim}[commandchars=\\\{\}]
tracel-ai/burn (49 files, \textcolor{checkgreen}{+1010}/\textcolor{black}{-945})
|-- crates/burn-autodiff/src/ops/int_tensor.rs
|-- crates/burn-autodiff/src/ops/tensor.rs
|-- crates/burn-backend/src/backend/ops/tensor.rs
|-- crates/burn-backend/src/backend/ops/int_tensor.rs
|-- crates/burn-backend/src/element/mod.rs
|-- crates/burn-backend/src/element/scalar.rs      \textcolor{checkgreen}{[NEW]}
|-- crates/burn-candle/src/ops/tensor.rs
|-- crates/burn-fusion/src/ops/int_tensor.rs
|-- crates/burn-router/src/ops/int_tensor.rs
\textcolor{gray}{|-- ... (40 more files)}
\end{Verbatim}
}

\noindent\textbf{Problem statement (first paragraph):}
\begin{quote}
\small\itshape
In the Burn deep learning framework, scalar arguments to tensor operations such as \texttt{add\_scalar}, \texttt{clamp\_min}, \texttt{equal\_elem}, \texttt{mask\_fill}, \texttt{powf\_scalar}, and similar functions currently use backend-specific element types (e.g., \texttt{FloatElem<B>}, \texttt{IntElem<B>}) or raw primitives (like \texttt{f32}, \texttt{i32}). This creates pervasive type-mismatch friction across the codebase: autodiff checkpointing must store and replay scalar values without a common runtime representation, fusion engines must serialize scalars into IR without knowing the originating tensor's concrete element type, and cross-backend routers must forward scalar literals through layers where the backend type is erased. Every new scalar-accepting operation requires duplicating conversion boilerplate (\texttt{.elem()} calls, explicit casts) across autodiff, fusion, router, and backend implementation files.

[\ldots]
\end{quote}

\subsection{Go: OpenListTeam/OpenList (47 files, 608 LOC)}
\label{app:case-go}

\noindent\textbf{Commit:} \url{https://github.com/OpenListTeam/OpenList/pull/1001}

\noindent\textbf{File tree (excerpt):}
{\small
\begin{Verbatim}[commandchars=\\\{\}]
OpenListTeam/OpenList (47 files, \textcolor{checkgreen}{+651}/\textcolor{black}{-375})
|-- drivers/115/driver.go
|-- drivers/115_open/upload.go
|-- drivers/123/upload.go
|-- drivers/alias/driver.go
|-- drivers/cloudreve/util.go
|-- internal/stream/stream.go
|-- internal/stream/util.go
|-- internal/stream/stream_test.go       \textcolor{checkgreen}{[NEW]}
|-- pkg/buffer/bytes.go                  \textcolor{checkgreen}{[NEW]}
|-- pkg/buffer/bytes_test.go             \textcolor{checkgreen}{[NEW]}
\textcolor{gray}{|-- ... (37 more files)}
\end{Verbatim}
}

\noindent\textbf{Problem statement (first paragraph):}
\begin{quote}
\small\itshape
During file upload operations across multiple storage drivers---including 115 Cloud, 123 Cloud, Google Drive, OneDrive, and others---the application encounters critical reliability issues rooted in stream handling, caching behavior, and underlying buffer management. These problems affect several distinct but interrelated subsystems.

[\ldots]
\end{quote}

\subsection{Python: google/langextract (30 files, 1{,}960 LOC)}
\label{app:case-python}

\noindent\textbf{Commit:} \url{https://github.com/google/langextract/pull/239}

\noindent\textbf{File tree (excerpt):}
{\small
\begin{Verbatim}[commandchars=\\\{\}]
google/langextract (30 files, \textcolor{checkgreen}{+2455}/\textcolor{black}{-619})
|-- examples/ollama/Dockerfile
|-- examples/ollama/demo_ollama.py       \textcolor{checkgreen}{[NEW]}
|-- examples/ollama/quickstart.py
|-- langextract/annotation.py
|-- langextract/core/base_model.py
|-- langextract/core/format_handler.py   \textcolor{checkgreen}{[NEW]}
|-- langextract/core/schema.py
|-- langextract/extraction.py
|-- langextract/resolver.py
|-- tests/format_handler_test.py         \textcolor{checkgreen}{[NEW]}
\textcolor{gray}{|-- ... (20 more files)}
\end{Verbatim}
}

\noindent\textbf{Problem statement (first paragraph):}
\begin{quote}
\small\itshape
The library supports extraction workflows across multiple language model providers---such as Ollama, Gemini, and others---each with distinct expectations for output structure. These include variations in serialization format (JSON vs. YAML), the presence of Markdown fence markers (e.g., \texttt{json} blocks), wrapper keys like ``extractions'', and attribute naming conventions. Currently, the logic governing these format decisions is distributed across prompt generation, model configuration, resolver parsing, and provider-specific implementations. This fragmentation creates subtle but persistent challenges during integration and maintenance.

[\ldots]
\end{quote}

\subsection{TypeScript: ant-design/ant-design (27 files, 97 LOC)}
\label{app:case-ts}

\noindent\textbf{Commit:} \url{https://github.com/ant-design/ant-design/pull/53739}

\noindent\textbf{File tree (excerpt):}
{\small
\begin{Verbatim}[commandchars=\\\{\}]
ant-design/ant-design (27 files, \textcolor{checkgreen}{+132}/\textcolor{black}{-48})
|-- .dumi/theme/builtins/ComponentTokenTable/index.tsx
|-- .dumi/theme/common/ComponentChangelog/ComponentChangelog.tsx
|-- components/avatar/AvatarGroup.tsx
|-- components/collapse/Collapse.tsx
|-- components/drawer/index.tsx
|-- components/dropdown/dropdown.tsx
|-- components/image/index.tsx
|-- components/modal/Modal.tsx
|-- components/tabs/index.tsx
|-- components/tooltip/index.tsx
\textcolor{gray}{|-- ... (17 more files)}
\end{Verbatim}
}

\noindent\textbf{Problem statement (first paragraph):}
\begin{quote}
\small\itshape
In the Ant Design component library, multiple components---including Modal, Drawer, Collapse, Tabs, Tooltip, Dropdown, and Image---provide functionality to unmount internal content when the component becomes hidden from view. This capability is critical for optimizing performance in complex applications, particularly when managing resource-intensive child elements or preserving clean component state between interactions. However, the current implementation suffers from significant API inconsistency across the component suite.

[\ldots]
\end{quote}

\end{document}